\documentclass{bmvc2k}

\newcount\Comments  % 1 suppresses notes to selves in text
\usepackage{color}
\definecolor{darkgreen}{rgb}{0,0.5,0}
\newcommand{\kibitz}[2]{\ifnum\Comments=0\textcolor{#1}{#2}\fi}

\usepackage{booktabs}
\usepackage{longtable}

\usepackage{listings}% http://ctan.org/pkg/listings
\usepackage{amssymb}

\usepackage{url}
\makeatletter
\g@addto@macro{\UrlBreaks}{\UrlOrds}
\makeatother

\newenvironment{myitem}
{ \begin{itemize}
    \setlength{\itemsep}{0pt}
    \setlength{\parskip}{0pt}
    \setlength{\parsep}{0pt}     }
{ \end{itemize}                  }

\title{The Interstate-24 3D Dataset: a new benchmark for 3D multi-camera vehicle tracking}

\addauthor{Derek Gloudemans}{derek.gloudemans@vanderbilt.edu}{1,2}
\addauthor{Gracie Gumm}{gracie.gumm@vanderbilt.edu}{1,2}
\addauthor{Yanbing Wang}{yanbing.wang@vanderbilt.edu}{1,2}
\addauthor{Will Barbour}{william.w.barbour@vanderbilt.edu}{1,2}
\addauthor{Daniel B. Work}{dan.work@vanderbilt.edu}{1,2}

\addinstitution{
Vanderbilt University \\
2201 West End Ave \\
Nashville, TN 37235
}

\addinstitution{
Vanderbilt University \\
Institute for Software Integrated Systems \\
1025 16th Ave S \\
Nashville, TN 37212
}

\runninghead{Gloudemans et al.}{The I-24 Multi-Camera 3D Tracking Dataset}

\begin{document}

\maketitle

\begin{abstract}
This work presents a novel video dataset recorded from overlapping highway traffic cameras along an urban interstate, enabling multi-camera 3D object tracking in a traffic monitoring context. Data is released from 3 scenes containing video from at least 16 cameras each, totaling 57 minutes in length. 877,000 3D bounding boxes and corresponding object tracklets are fully and accurately annotated for each camera field of view and are combined into a spatially and temporally continuous set of vehicle trajectories for each scene. Lastly, existing algorithms are combined to benchmark a number of 3D multi-camera tracking pipelines on the dataset, with results indicating that the dataset is challenging due to the difficulty of matching objects travelling at high speeds across cameras and  heavy object occlusion, potentially for hundreds of frames, during congested traffic. This work aims to enable the development of accurate and automatic vehicle trajectory extraction algorithms, which will play a vital role in understanding impacts of autonomous vehicle technologies on the safety and efficiency of traffic.
\end{abstract}

 % A thorough error analysis is performed on the data, and sources of error arising from field deployments of IP cameras are corrected to improve the quality of the data.

%-------------------------------------------------------------------------
\section{Introduction}

In recent years, 3D detection and tracking datasets in the autonomous vehicle domain have led to marked advancements in perception and planning algorithms and AV technology more generally \cite{geiger2013vision,caesar2020nuscenes,sun2020scalability}. But designing autonomous technologies from an ego-vehicle perspective alone is not enough. Studies have shown that control algorithms designed for an individual vehicle's objectives can cause rippling instabilities in traffic \cite{gunter2020commercially}, while controllers designed with global traffic objectives in mind can significantly reduce congestion \cite{stern2018dissipation,wu2017flow}. 

Automatic traffic monitoring offers a tremendous but under-exploited opportunity to address this issue. Computer vision research has progressed sufficiently in other fields such that efficient algorithms for traffic monitoring at scale likely exist, and state and federal transportation agencies maintain camera networks with tens of thousands of cameras nationally; increasingly ubiquitous edge sensing devices only add to the number of potentially useful traffic cameras.  Moreover, in several cases, multi-camera systems have been deployed at considerable scale specifically to study the effects of \textit{intelligent transportation systems} (ITS) and AVs on traffic \cite{tgsim,gloudemans202324,von2021creating,seo2020evaluation,michigan-corridor}. Similarly, work on autonomous management of city-scale traffic will benefit immensely from the ability to track vehicle movements precisely (often requiring 3D detection) across many cameras \cite{tang2019cityflow}. It yet remains to be explored whether existing algorithms can achieve tracking performance suitable for fine-grained traffic analysis (i.e. HOTA above 75\% and over 95\% mostly tracked objects), where small localization errors or a single ID switch can be damaging in understanding a scenario \cite{coifman2017critical}.

We seek to enable research on precise vehicle tracking in the traffic monitoring context, with emphasis on the challenges of multi-camera tracking faced in systems such as \cite{tgsim,gloudemans202324,von2021creating,seo2020evaluation,michigan-corridor}. Work in this field has been slowed by a lack of 3D multi-camera tracking data; this work addresses this shortage to enable development and evaluation of tracking methods to meet the needs of the next generation of intelligent traffic systems and AV research.

\begin{figure}
    \centering
    \includegraphics[width = \columnwidth]{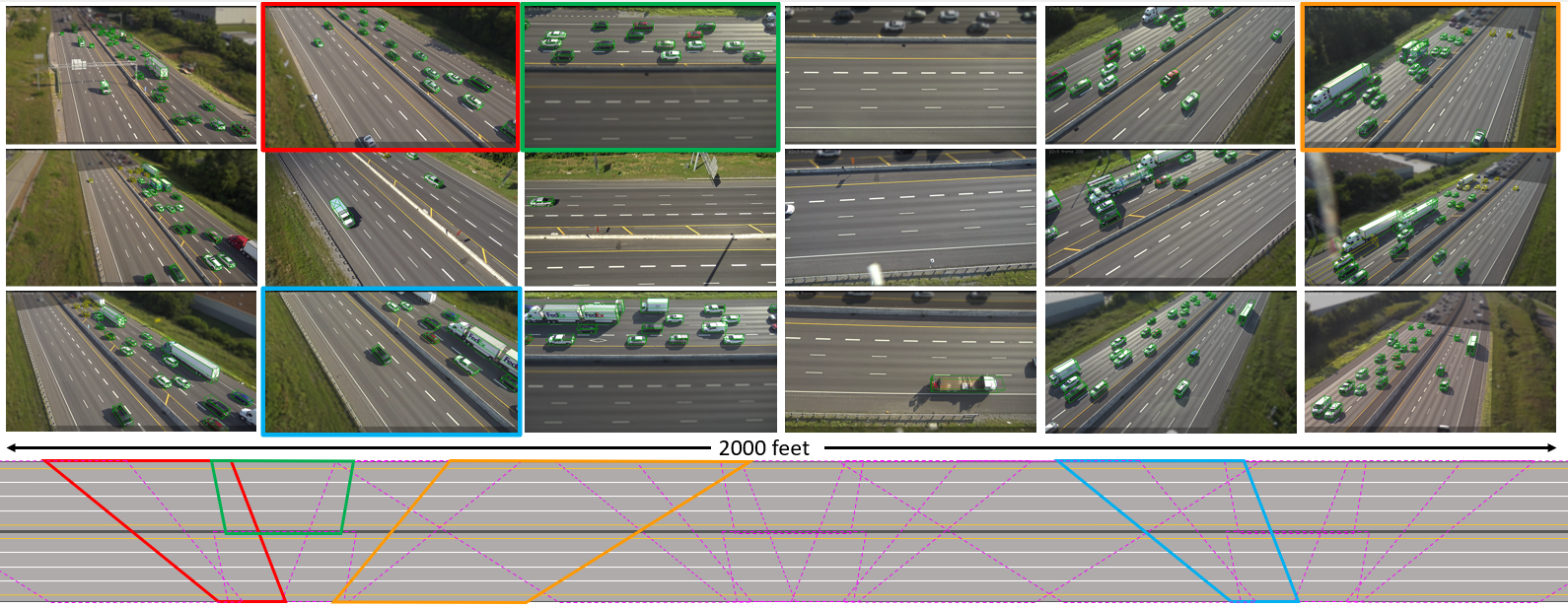}
    \caption{Example annotated (green boxes) frames from each camera field of view for one scene of the I24-3D Dataset. The approximate field of view for each camera is shown on the overhead roadway diagram below (some cameras shown in unique colors as examples). Regions outside of the considered field of view for each camera are blurred for this visualization. Cameras provide coverage of 2000 feet of Interstate-24 near Nashville, TN.}
    \label{fig:intro}
    \vspace{-0.25in}
\end{figure}

\textbf{The primary contribution of this work is the introduction of a novel dataset suitable for multi-camera tracking, consisting of 877,000 3D vehicle bounding boxes annotated across 16-17 cameras with dense viewpoints covering 2000 feet of interstate roadway near Nashville, TN. }The \textit{Interstate-24-3D Dataset} (I24-3D) introduced in this work is comprised of 3 \textit{scenes} (sets of videos recorded at the same time from different cameras), recorded at 4K resolution and 30 frames per second. Vehicle 3D bounding boxes are annotated by hand for 720 unique vehicles. I24-3D is the first 3D multiple-camera dataset in a traffic monitoring context and tracks objects across a larger set of cameras than any other multi-camera tracking dataset.\textbf{ The secondary contribution of this work is the benchmarking of a number of  existing algorithm combinations to assess the difficulty of 3D multi-camera tracking on this dataset,} with results (best performance of 44.8\% HOTA and 62\% mostly tracked objects) showing that the implemented methods achieve good performance but do not produce data suitable for fine-grained traffic analysis.

The rest of this paper is organized as follows: Section \ref{sec:lit} reviews the most analogous existing datasets. Section \ref{sec:overview} describes the data and annotations included in I24-3D. Section \ref{sec:experiments} provides details of benchmarking experiments using the dataset, and Section \ref{sec:results} describes the results. Additional details on the dataset including example video links, annotation details, file format and accuracy metrics, timestamp synchronization efforts, additional experimental settings and implementation details, unabridged results, and privacy considerations are included in Appendices I through VII.

\section{Related Work}
\label{sec:lit}
\textbf{Vehicle Trajectories:} Traffic trajectory data consists of vehicle positional data for each vehicle within a traffic stream. Such data is required for myriad traffic analysis and modeling applications, yet sources are limited: the NGSIM dataset \cite{alexiadis2004next}, known to contain large vehicle positional errors \cite{coifman2017critical}, and the HighD dataset \cite{krajewski2018highd} are the two main datasets and are limited in time and space. Some additional works utilize sensor-equipped vehicles \cite{hankey2016description} or GPS data \cite{liu2012calibrating,shi2009automatic} to collect individual vehicle trajectories, but do not provide data for the majority of vehicles. Such a shortage of traffic trajectory data requires that researchers rely on models to approximate human driving behavior  \cite{lee2009environmental,sun2019modelling,lopez2018microscopic}. Recent efforts have sought to provide additional trajectory data using video data and computer vision \cite{tgsim,seo2020evaluation,von2021creating}. 

\textbf{Trajectory generation methods:} To address the trajectory data shortage, several methods have been proposed to automatically extract vehicle trajectory data from existing traffic cameras. \cite{dubska2014automatic} proposes a method to detect vehicle 3D rectangular prism bounding boxes using background subtraction and blob segmentation, relying on automatic parameter extraction of  scene homography proposed in \cite{dubska2014fully}. The results are validated on derivative data products (vehicle speeds and lane positions). \cite{ren2018learning} uses 2D object detectors to roughly estimate vehicle positions on the road plane, and \cite{subedi2019development} uses ground plane projection of vehicle pixels from multiple cameras to estimate the vehicle's position, validating with turning movement counts. Other solutions use re-identification of 2D tracked objects, without addressing 2D annotation position ambiguity \cite{tang2018single,chen2019multi}. Lastly, 3D vehicle multiple object detection and tracking methods such as \cite{zhou2021monocular, xiang2015data, li2020rtm3d,chabot2017deep,sochor2018boxcars} can also be applied to produce vehicle trajectories. A few works have addressed the multi-camera 3D tracking problem, either relying on fusing detections in a shared space 
 \cite{caesar2020nuscenes,strigel2013vehicle,luna2022online} or else fusing tracklets from individual cameras after tracking \cite{he2020multi,wang2022automatic}. 

\textbf{Multiple object tracking datasets:} The task of single camera 2D \textit{multiple object tracking} (MOT) is well-studied in varied contexts, including pedestrian and vehicle tracking from stationary and moving cameras (MOT16) \cite{milan2016mot16}, tracking from drone footage (VISDRONE) \cite{zhu2018visdrone}, and traffic monitoring (UA-DETRAC) \cite{wen2020ua}. 3D single camera (or stereo camera for depth) MOT is also well-addressed within the domain of \textit{autonomous vehicle} (AV) or ego-vehicle data (KITTI, Waymo OpenDrive, and NuScenes) \cite{geiger2013vision,sun2020scalability,caesar2020nuscenes}. Data annotation in this context is aided by rich LIDAR data from on-vehicle sensors. Rich 3D data in the traffic monitoring (overhead traffic camera) domain is sparse, in part because LIDAR sensors are not collocated with cameras to aid in annotation. Only the BoxCars116k dataset \cite{sochor2018boxcars}  provides 3D monocular bounding boxes. Thus, research on 3D vehicle tracking from overhead cameras must use simulated or partially synthetic data \cite{buhet2019conditional,zhu2021monocular,miao2021robust}.

\textbf{Multi-camera tracking datasets: } \textit{Multiple camera multiple object tracking} are few in number and are mostly in the context of pedestrian tracking. The Duke-MTMC dataset associated 2D object tracklets for pedestrians across 8 cameras \cite{gou2017dukemtmc4reid}, and the PETS dataset \cite{ferryman2009pets2009}, EPFL Terrace \cite{liu2017multi}, EPFL-RLC \cite{chavdarova2017deep}, and WILDTRACK \cite{chavdarova2018wildtrack} synchronize up to 7 cameras for pedestrian multi-camera tracking \cite{liu2017multi}, These datasets provide annotations in a unified ground plane, with pedestrians represented as points \cite{chavdarova2018wildtrack} or grid cell occupants \cite{chavdarova2017deep} on the ground plane. In a vehicle context, the CityFlow dataset \cite{tang2019cityflow} associates 2D MOT data in a traffic monitoring context across multiple cameras throughout a city, with an average of 4 cameras covering scenes, but object dimensions and space are not modeled. NuScenes contains multiple frontal, side and rear-facing, frame-capture synchronized cameras enabling 3D multiple-camera tracking in an AV context.  To the best of our knowledge, no multiple-camera traffic monitoring dataset with 3D object tracking annotations exists. 

\section{The I24-3D Dataset}
\label{sec:overview}
This section introduces the I24-3D Dataset, detailing the location of the cameras, describing the annotations, vehicle classes, and suitable uses, and providing annotation quality metrics.

\subsection{Overview}
The I24-3D Dataset consists of 3 \textit{scenes}, or collections of video data recorded simultaneously from 16-17 cameras, densely covering a section of roughly 2000 feet of roadway. Each scene is 60-90 seconds long, recorded at 4K resolution and 30 frames per second, and features manually annotated 3D bounding boxes on every vehicle visible within the field of view of each camera suitable for vehicle re-identification, 3D object detection, tracking, and multi-camera tracking tasks (see Table \ref{tab:data-summary}.) Over 275 person hours were spent annotating the data. A full description of the dataset file structure and format is included in Appendix I, and example videos are included in supplementary material for review.

\begin{table}[b]
%\vspace{-0.1in}
\centering
\renewcommand{\arraystretch}{0.8}
\begin{tabular}{@{}lc@{\hskip 0.1in}c@{\hskip 0.1in}c@{\hskip 0.1in}c@{\hskip 0.1in}c@{\hskip 0.1in}c@{\hskip 0.1in}l@{}l}
\toprule
Scene & Time (s) & Cameras & Frames & Boxes & IDs & VMT & Description\\ \midrule
1 & 90 & 17 & 45900 & 291k & 324 & 118 & Free-flow traffic\\
2 & 60 & 16 & 30600 & 146k & 114 & 24.4 & Slow traffic, snow conditions\\ 
3 & 60 & 16 & 28800 & 440k & 282 & 67.0 & Congested traffic \\ \midrule
Total &  210 & - & 105300  & 877k & 720 & 209 & -\\ \bottomrule
\end{tabular}
\caption{Summary of scene data for I24-3D dataset. \textit{Time} indicates the total global duration of a scene (each video segment for the scene has that duration). \textit{Frame} count is aggregated across all cameras in the scene, \textit{cameras} indicates the number of active cameras for the scene. \textit{Boxes} indicates number of 3D bounding boxes, \textit{IDs} indicates unique vehicle trajectories.}
\vspace{-0.15in}
\label{tab:data-summary}
\end{table}

\subsection{Location}
I24-3D was recorded using I-24 MOTION \cite{gloudemans202324}, an open-road testbed along Interstate 24 near Nashville, Tennessee. The utilized portion of this testbed contains 18 cameras mounted on three 110-foot tall roadside poles, spaced at roughly 500 feet and covering an approximately 2000 foot field of view on the interstate \cite{barbour2021interstate}. (Due to periodic camera outages, each scene contains footage from only 16-17 cameras).

% \begin{figure}
%     \centering
%     \includegraphics[width = \textwidth]{images/scenes.png}
%     \caption{Example camera frames from each I24-3D scene.} 
%     \label{fig:scenes}
%     \vspace{-0.40in}
% \end{figure}

\subsection{Annotation Description}
Annotations are provided in a roadway-aligned coordinate plane, where x-coordinate indicates distance along the roadway and y-coordinate indicates lateral (lane) position, of the bottom center rear of the vehicle. For each direction of travel in each camera field of view, a \textit{homography} relates the roadway coordinate system to the pixel coordinates of the field of view. We rely on standard perspective transforms \cite{hartley2003multiple} for this conversion (see Appendix II), assuming the roadway visible in each field of view can be reasonably represented by a flat plane with a relevant \textit{field of view} (FOV) comprising most of each image (masks are provided for regions falling outside of the FOV for each camera). All distances are given feet, as the geometry of the roadway is laid out in feet (e.g. lanes are 12 feet wide).

\begin{figure}[t]
    \vspace{-0.05in}
    \centering
    \includegraphics[width = \columnwidth]{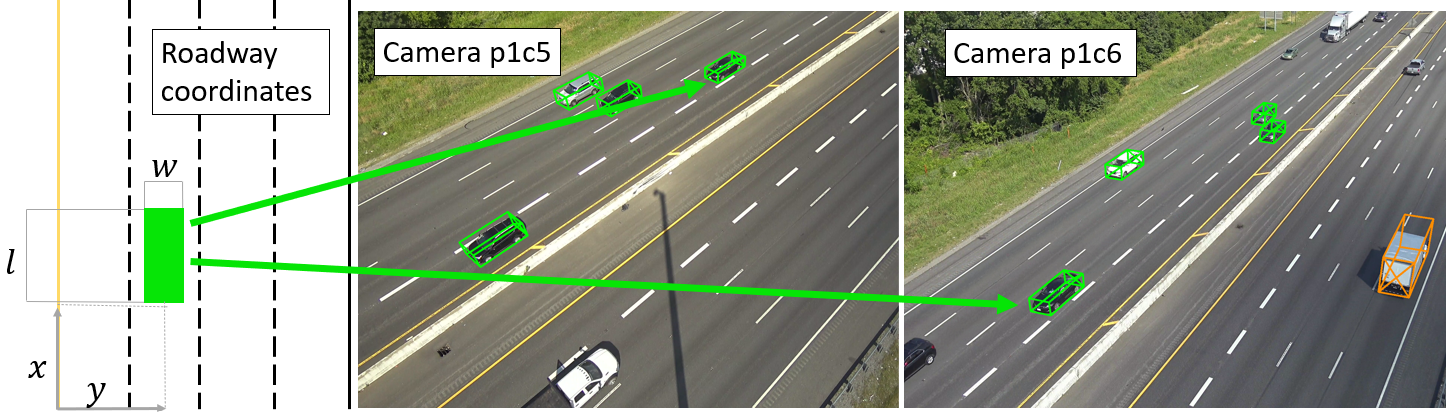}
    \caption{Example single annotation. The annotation is stored in roadway coordinates (left) but can be projected into cameras 5 and 6 on pole 1 (p1c5 and p1c6).} 
    \label{fig:state}
    \vspace{-0.3in}
\end{figure}

A single vehicle 3D bounding box annotation includes \textit{vehicle class}, unique \textit{vehicle ID}, bounding box \textit{length}, \textit{width}, and \textit{height} (fixed for all annotations for a single vehicle), \textit{vehicle roadway position}, \textit{originating camera}, \textit{timestamp}, and \textit{frame index}. This information is sufficient to losslessly project the annotation into the originating camera, or into any other camera in which it is visible. Figure \ref{fig:state} shows an example.  \textbf{Object localization is precise, with 1.24 ft average positional error between annotations of the same vehicle labeled in multiple cameras, and 0.5 ft average dimensional error}. (See Appendix III and IV.)

\subsection{Vehicle Classes}
\begin{figure*}[h]
    \centering
    \includegraphics[width = \textwidth]{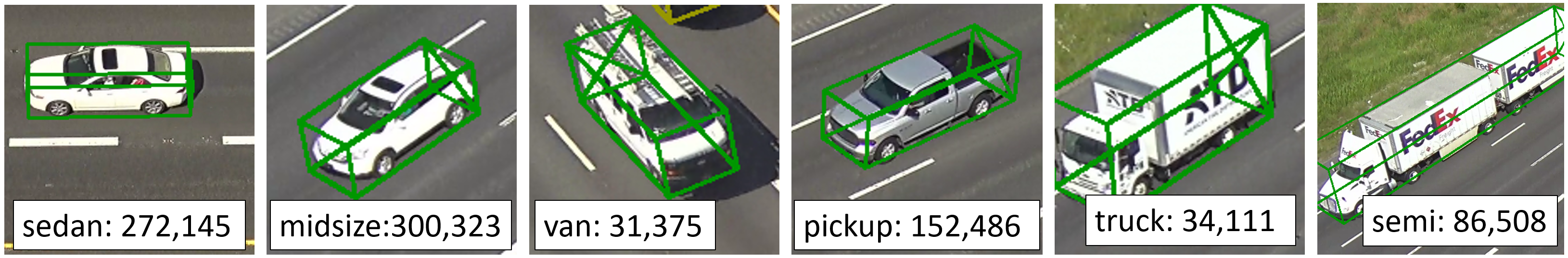}
    \caption{Example vehicles and vehicle class annotation counts for the I24-3D dataset.}
    \label{fig:classes}
    \vspace{-0.25in}
\end{figure*}

Vehicles are classified into six classes: \textit{sedan}, \textit{midsize} (minivan, SUV or compact SUV), \textit{van}, \textit{pickup}, \textit{semi} (tractor-trailer), or \textit{truck}). Figure \ref{fig:classes} depicts example annotations for each class as well as the total number of annotations for each class.  We make one additional distinction: vehicles other than semis that tow trailers are classified with the towing vehicle's class, but bounding boxes are drawn to include the trailer. This choice reflects that a vehicle and trailer behave as a single semi-rigid body. Vehicle IDs with trailers include: Scene 1: [288, 133, 7, 138, 43, 270, 245, 216], Scene 3: [225, 105, 15, 148, 247, 219].

\subsection{Dataset Uses and Comparison}
The I24-3D dataset provides annotations of sufficient richness for a variety of canonical computer vision problems, including object reidentification, 2D and 3D detection and tracking. Most notably, multiple videos from a single scene can be used for multiple-camera tracking tasks, and the presence of 3D labels in this dataset enables explicit modeling of a shared 3D space for object tracking. Table \ref{tab:datasets} provides a comparison of the suitable uses of the I24-3D dataset and the most similar existing datasets. Notably, I24-3D is the only dataset in a traffic monitoring context that allows for 3D multi-camera tracking.

% Agg Table
\begin{table*}[t]
\setlength{\tabcolsep}{3pt} % Default value: 6pt
\renewcommand{\arraystretch}{0.85} % Default value: 1
\centering
\begin{tabular}{@{}lccc@{\hskip 0.1in}c@{\hskip 0.05in}ccccccccc@{}}
\toprule
Dataset & Resolution & \multicolumn{2}{l}{Detection} & \multicolumn{2}{l}{MOT} & \multicolumn{2}{l}{MCT} & Boxes & Frames & Cameras  \\   \cmidrule{3-8}

  & &  2D & 3D & 2D & 3D & 2D & 3D &  \\ \toprule
WILDTRACK \cite{chavdarova2018wildtrack}  & 1920$\times$1080   & \checkmark &   &  \checkmark &   &  \checkmark & \checkmark & 38k & 61k & 7   \\ 
KITTI \cite{geiger2013vision}  & 1382$\times$512  &    \checkmark &  \checkmark &  \checkmark &  \checkmark &   &   & 200k & 15k & 1 \\ 
NuScenes \cite{caesar2020nuscenes}  & 1600$\times$900 &  \checkmark &  \checkmark &  \checkmark &  \checkmark &  \checkmark &  \checkmark & 12M & 40k & 6 \\ \midrule 
BoxCars116k \cite{sochor2018boxcars} & \textit{varies}    &  \checkmark &  \checkmark &   &   &   &  & 116k & 116k & 1   \\
UA-DETRAC \cite{wen2020ua} &  1920$\times$1080 &    \checkmark &   &  \checkmark &   &   &  & 1.2M & 140k & 1   \\
CityFlow \cite{tang2019cityflow} &  960$\times$540    &  \checkmark &   &  \checkmark &   &  \checkmark &  & 229k & 117k & 4 \\ 
\textbf{I24-3D (Ours)} & 3840$\times$2160 & \checkmark &  \checkmark  &  \checkmark &  \checkmark &  \checkmark &  \checkmark  & 877k & 105k & 16-17 \\ \bottomrule
\end{tabular}
\caption{Suitable uses and metrics for comparable MOT and 3D vehicle detection datasets, grouped by traffic monitoring (bottom) and other contexts (top). \textit{MOT} indicates multiple object tracking (in 2D or 3D), and \textit{MCT} indicates multiple camera tracking (with 3D tracking requiring a unified tracking space). \textit{Boxes} indicates the total number of monocular bounding box view annotations, \textit{Frames} indicates the total number of annotated frames in the dataset, \textit{Cameras} indicates the number of camera views in a single scene.}
\label{tab:datasets} 
\vspace{-0.2in}
\end{table*}

\section{Benchmarking Experiments}
\label{sec:experiments}

To provide an initial gauge of tracking difficulty and existing algorithm performance on I24-3D, we benchmark a set of tracking methods on this dataset. Experimental protocol, metrics for evaluation, and implemented algorithms are briefly described in this section.

\subsection{Experimental Protocol}
Each scene is split into temporally contiguous training and validation partitions (the first 80\% and the last 20\% of each scene, respectively). Training is performed exclusively using the training partition. All training is performed locally on RTX6000 GPUs, and detection models are trained until convergence. During tracking we maintain tight 1/60th second synchronization between each video using corrected frame timestamps (see Appendix III), skipping frames as necessary to maintain a 15 Hz nominal frame rate.

For tracking evaluation, we find a best-fit 3rd order polynomial spline for each ground truth vehicle to obtain a continuous object representation in roadway coordinates. Predicted vehicle trajectories are compared against boxes sampled from the best-fit spline for each object. We linearly interpolate between the spline-sampled boxes and the tracker-output predictions at 30Hz to produce object sets at the same discrete times. Additional experimental details are given in Appendix V. 

\subsection{Metrics}
We compare tracker performance using the clearMOT metrics \cite{bernardin2008evaluating}, the MT/ML metrics used in \cite{wu2006tracking}, and HOTA \cite{luiten2021hota}. We also consider the percentage of ground truth ($GT\%$) and predicted objects ($Pred\%$) matched to at least one predicted or ground truth object, respectively). To account for time-synchronization errors, we use a requisite 30\% 2D-footprint IOU threshold between predicted and ground truth objects.

\subsection{Algorithms Implemented}
A variety of multi-camera 3D MOT pipelines are assembled, each requiring 3 algorithmic components: i.) a 3D object detector, ii.) an object tracker / association method, and iii.) a method for combining objects across cameras. We briefly describe algorithms implemented for each stage (implementation and parameter details can can be found in Appendix V).

\noindent\textbf{3D Detectors:}
\vspace{-0.05in}
\begin{myitem}
    \item \textbf{Monocular 3D Detector (Single3D)} - a Retinanet model with Resnet34-FPN backbone \cite{lin2017focal}. The formulation is camera-agnostic (as training a separate model for each camera FOV is infeasible both from data scarcity and scalability standpoints.) \textit{Average Precision} (AP) scores for this detector: $AP_{30} = 0.718$, $AP_{50} = 0.598$, $AP_{70} = 0.254$. (See Appendix V for experimental details.)
    \item \textbf{Monocular 3D Multi-frame Detector (Dual3D)} - Inspired by recent works utilizing multiple frames for detection and tracking \cite{zhou2020tracking}, we add the previous frame as detection input. AP scores for this detector: $AP_{30} = 0.810$, $AP_{50} = 0.714$, $AP_{70} = 0.572$.
    \item \textbf{Monocular 3D Crop Detector (CBT)} - as described in \cite{gloudemans2021vehicle}, we train a Retinanet Model with Resnet34-FPN backbone for detecting objects in cropped portions of full frames. AP  scores for this detector: $AP_{30} = 0.767$, $AP_{50} = 0.700$, $AP_{70} = 0.464$.
    \item \textbf{Ground Truth Detections (GT)} - perfect ground-truth detections.
\end{myitem}

\noindent\textbf{Object Trackers:}
\vspace{-0.05in}
\begin{myitem}
    \item \textbf{Kalman-Filter IOU Tracker (KIOU)} - as described in \cite{bochinski2017high}. We utilize a contant velocity roadway-coordinate Kalman filter for object position prediction.
    \item \textbf{ByteTracker (Byte)} - noting this tracker's state of the art performance on the MOTChallenge benchmarks \cite{milan2016mot16}, we utilize the two-stage association method described in \cite{zhang2022bytetrack}, using IOU as both primary and secondary matching criterion and utilizing a Kalman filter as suggested by authors.
    \item \textbf{Crop-based Tracking (CBT)} - as proposed in \cite{gloudemans2021vehicle}, detection on some frames is performed by re-detecting priors in cropped subsets of the overall frame, and object associations are implicit for these frames. 
    \item \textbf{Ground Truth Single Camera Tracklets} - perfect single-camera tracklets.
\end{myitem}

\noindent\textbf{Cross-Camera Rectification Methods:}
\vspace{-0.05in}
\begin{myitem}
    \item \textbf{Detection Fusion (DF)} - as preferred in the AV context \cite{caesar2020nuscenes}, detections from all cameras are combined online in roadway coordinates and non-maximal supression with a stringent 1\% IOU threshold utilized to eliminate overlapping detections. 
    \item \textbf{Trajectory Fusion (TF)}- as proposed in \cite{wang2022automatic}, single camera tracklets are compared for spatio-temporal overlap offline, stitched together when a matching criteria is met, and refined to optimally describe the observed set of tracked object positions.
    \item \textbf{None} - as a baseline, object tracklets from each camera are output with no fusion.
    \item \textbf{Both (DF+TF)} - Tracking uses detection fusion, and a subsequent trajectory stitching step is performed to deal with remaining object fragmentations.
\end{myitem}

\section{Results}
\label{sec:results}

\begin{table}[t]
\setlength{\tabcolsep}{2.2pt} % Default value: 6pt
\renewcommand{\arraystretch}{0.76}
\begin{tabular}{@{}lccc|ccccccccc@{}} \toprule
\textbf{Detector} & \textbf{Tracker} & \textbf{DF} & \textbf{TF} & \textbf{HOTA} & \textbf{MOTA} & \textbf{Rec} & \textbf{Prec} & \textbf{GT\%} & \textbf{Pred\%} & \textbf{MT} & \textbf{ML} & \textbf{Sw/GT} \\ \toprule

Crop & Byte & \checkmark & \checkmark & 23.6 & 21.3 & 53.4 & 64.0 & 90.5 & 72.9 & 25.6 & 25.0 & 1.1 \\
Crop & KIOU & \checkmark & \checkmark & 24.6 & 21.4 & 54.4 & 64.2 & 90.5 & 71.2 & 27.6 & 22.3 & 1.1 \\
Dual3D & Byte & \checkmark & \checkmark & 30.9 & 50.0 & 65.6 & 81.9 & 90.6 & 93.4 & 35.9 & 15.0 & 0.9 \\
Dual3D & KIOU & \checkmark & \checkmark & 39.7 & 71.6 & 76.5 & 93.7 & 91.5 & 95.3 & 52.5 & 10.4 & 0.7 \\
Single3D & Byte & \checkmark & \checkmark & 27.5 & 49.3 & 62.8 & 83.9 & 92.1 & 91.8 & 29.7 & 15.4 & 0.9 \\
Single3D & KIOU & \checkmark & \checkmark & 39.9 & 71.6 & 76.3 & 94.1 & 93.4 & \textbf{95.6} & 51.6 & 8.8 & 0.7 \\ \midrule
Crop & Byte &  & \checkmark & 15.2 & -16.5 & 43.5 & 47.0 & 90.4 & 59.8 & 14.5 & 32.2 & 1.5 \\
Crop & KIOU &  & \checkmark & 20.7 & -2.1 & 51.6 & 52.9 & 90.2 & 53.7 & 25.5 & 26.4 & 1.5 \\
Dual3D & Byte &  & \checkmark & 38.7 & 75.0 & 80.2 & 93.8 & 93.0 & 93.6 & 59.0 & 8.0 & 0.8 \\
Dual3D & KIOU &  & \checkmark & \textbf{44.8} & 77.0 & \textbf{83.0} & 93.2 & 91.7 & 92.3 & \textbf{63.8} & 8.8 & \textbf{0.5} \\
Single3D & Byte &  & \checkmark & 38.2 & 72.6 & 80.6 & 90.9 & \textbf{94.9} & 92.5 & 58.7 & \textbf{ 4.7} & 1.1 \\
Single3D & KIOU &  & \checkmark & \textbf{44.8} & \textbf{77.1} & \textbf{83.0} & 93.4 & 93.3 & 91.3 & 62.2 & 7.8 & \textbf{0.5} \\ \midrule \midrule
Crop & Byte & \checkmark &  & 19.2 & 34.7 & 58.3 & 72.8 & 91.9 & 81.4 & 27.1 & 16.6 & 2.4 \\
Crop & KIOU & \checkmark &  & 19.3 & 32.1 & 57.9 & 71.4 & 91.9 & 79.7 & 25.9 & 17.3 & 2.4 \\
Dual3D & Byte & \checkmark &  & 20.9 & 60.2 & 64.2 & 94.8 & 92.7 & 94.7 & 29.5 & 8.7 & 2.8 \\
Dual3D & KIOU & \checkmark &  & 21.1 & 60.4 & 64.0 & 95.2 & 92.6 & 94.7 & 29.7 & 8.9 & 2.7 \\
Single3D & Byte & \checkmark &  & 21.3 & 60.3 & 63.9 & 95.3 & 93.8 & 94.2 & 27.1 & 7.5 & 2.6 \\
Single3D & KIOU & \checkmark &  & 21.4 & 60.3 & 63.7 & \textbf{95.5} & 94.2 & 93.9 & 26.5 & 7.5 & 2.6 \\ \midrule
Crop & Byte &  &  & 17.6 & 18.7 & 59.8 & 64.0 & 91.9 & 73.0 & 30.4 & 15.1 & 3.0 \\
Crop & KIOU &  &  & 16.9 & 10.8 & 57.5 & 60.0 & 91.9 & 66.0 & 28.2 & 18.5 & 3.2 \\
Dual3D & Byte &  &  & 15.0 & 55.1 & 72.8 & 81.7 & 93.2 & 87.5 & 42.5 & 6.8 & 7.3 \\
Dual3D & KIOU &  &  & 15.1 & 55.6 & 72.7 & 82.2 & 93.1 & 87.8 & 42.3 & 7.0 & 7.3 \\
Single3D & Byte &  &  & 15.1 & 54.0 & 72.3 & 80.8 & 94.3 & 85.9 & 40.5 & 5.8 & 7.2 \\
Single3D & KIOU &  &  & 15.2 & 54.4 & 72.2 & 81.3 & 94.5 & 86.1 & 39.4 & 5.6 & 7.1 \\ \bottomrule

\end{tabular}
\caption{Tracking results for each multi-camera tracking pipeline. \textbf{Sw/GT} indicates object ID switches per ground truth object. Best result for each metric shown in bold.}
\label{tab:result}
\vspace{-0.1 in}
\end{table}

Table \ref{tab:result} reports results for each of the above implemented pipelines. The best performing pipeline combines Dual3D detection with KIOU tracking and trajectory fusion (HOTA 44.8\%). In general, trajectory fusion alone performs best (across otherwise equal run settings) and no cross-camera rectification strategy (baseline) performs worst. While relatively high MOTA scores are achievable at a low 0.3 IOU threshold (77.1\% maximum), HOTA scores are still relatively low when compared to top performing algorithms on MOTchallenge and KITTI \cite{milan2016mot16, geiger2013vision}. This is primarily driven by relatively low localization accuracy, especially for fast moving vehicles (where a 1-frame timing error results in dropping below a 70\% threshold for localization accuracy for an otherwise perfect detection.) See Appendix VI for an example HOTA plot at varying localization thresholds.

\textbf{Even the best pipelines miss 5\% of ground truth objects entirely (GT\%), and track only 64\% of objects for 80\% of overall duration (MT).} This result demonstrates the difficulty of tracking most or all of the vehicles in a traffic scene at the level of granularity and completeness necessary for in-depth traffic analysis. Even utilizing ground truth detections or single camera tracklets cannot fully mitigate these failures. For brevity, pipelines utilizing ground truth inputs are included in Appendix VI; the best-performing pipeline utilizing ground truth detections achieves HOTA 59.6\%, and the best-performing pipeline utilizing ground-truth single-camera tracklets achieves HOTA 61.6\%. This indicates that the cross-camera tracklet rectification problem is difficult even with great single-camera tracklets.  

\begin{table}[b]

\centering
\setlength{\tabcolsep}{4pt} % Default value: 6pt
\renewcommand{\arraystretch}{0.76}

\begin{tabular}{@{}l|cccccccccc@{}} \toprule
 \textbf{Scene} & \textbf{HOTA} & \textbf{MOTA} & \textbf{MOTP} & \textbf{Rec} & \textbf{Prec} & \textbf{GT\%} & \textbf{Pred\%} & \textbf{MT} & \textbf{ML} & \textbf{Sw/GT} \\ \toprule
% 1 & \textbf{57.5} & \textbf{89.3} & 50.4 & \textbf{92.9} & \textbf{96.3} & \textbf{96.6} & \textbf{97.9} & \textbf{87.0} & \textbf{1.6} & \textbf{0.0} \\
% 2 & 47.7 & 78.2 & \textbf{86.2} & 86.4 & 91.5 & 93.0 & 79.8 & 64.0 & 7.0 & 0.5 \\
% 3 & 29.1 & 63.8 & 82.3 & 69.7 & 92.3 & 90.4 & 96.1 & 35.6 & 14.9 & 1.0 \\ \midrule
% \textit{avg} & 44.8 & 77.1 & 73.0 & 83.0 & 93.4 & 93.3 & 91.3 & 62.2 & 7.8 & 0.5 \\ \bottomrule

1 & \textbf{58.5} & \textbf{89.7} & 69.2 & \textbf{92.9} & \textbf{96.7} & \textbf{95.3} & \textbf{98.4} & \textbf{86.3} & \textbf{2.2} & \textbf{0.02} \\
2 & 46.9 & 77.7 & \textbf{74.5} & 86.2 & 91.1 & 90.4 & 82.4 & 64.0 & 9.6 & 0.49 \\
3 & 29.1 & 63.5 & 64.8 & 69.9 & 91.7 & 89.3 & 96.1 & 40.9 & 14.6 & 1.05 \\ \midrule
\textit{avg} & 44.8 & 77.0 & 69.5 & 83.0 & 93.2 & 91.7 & 92.3 & 63.8 & 8.8 & 0.52 \\ \bottomrule

\end{tabular}
\caption{Tracking results for Dual3D + KIOU + TF for each scene. Best score for each metric shown in bold, generally suggesting an easier scene.}
 \label{tab:per-scene}
 \vspace{-0.1in}
\end{table}

Table \ref{tab:per-scene} reports results for the best pipeline per scene. Scene 1 is easiest across a variety of metrics, with Scene 2 being easier on MOTP (slow-moving objects due to snowy conditions minimizes localization inaccuracies). Per-scene results for all methods are included in Appendix VI. Figure \ref{fig:trajectories} shows the best performing pipeline's outputs evaluated against ground truth object annotations for Scene 3. Lanes farther from cameras and with high object densities have a much higher rate of false negatives (e.g. westbound (WB) lane 4). Slow-moving, un-occluded objects (e.g. WB Lane 1) are tracked relatively accurately. Faster moving objects (e.g. EB Lane 2) are often tracked, but not accurately enough to surpass the IOU threshold requirement. Results on Scene 3 demonstrate the difficulty of tracking all objects in dense stop-and-go traffic, when many objects are occluded for long periods of time.

\begin{figure} [ht]
    \centering
    \includegraphics[width=\textwidth]{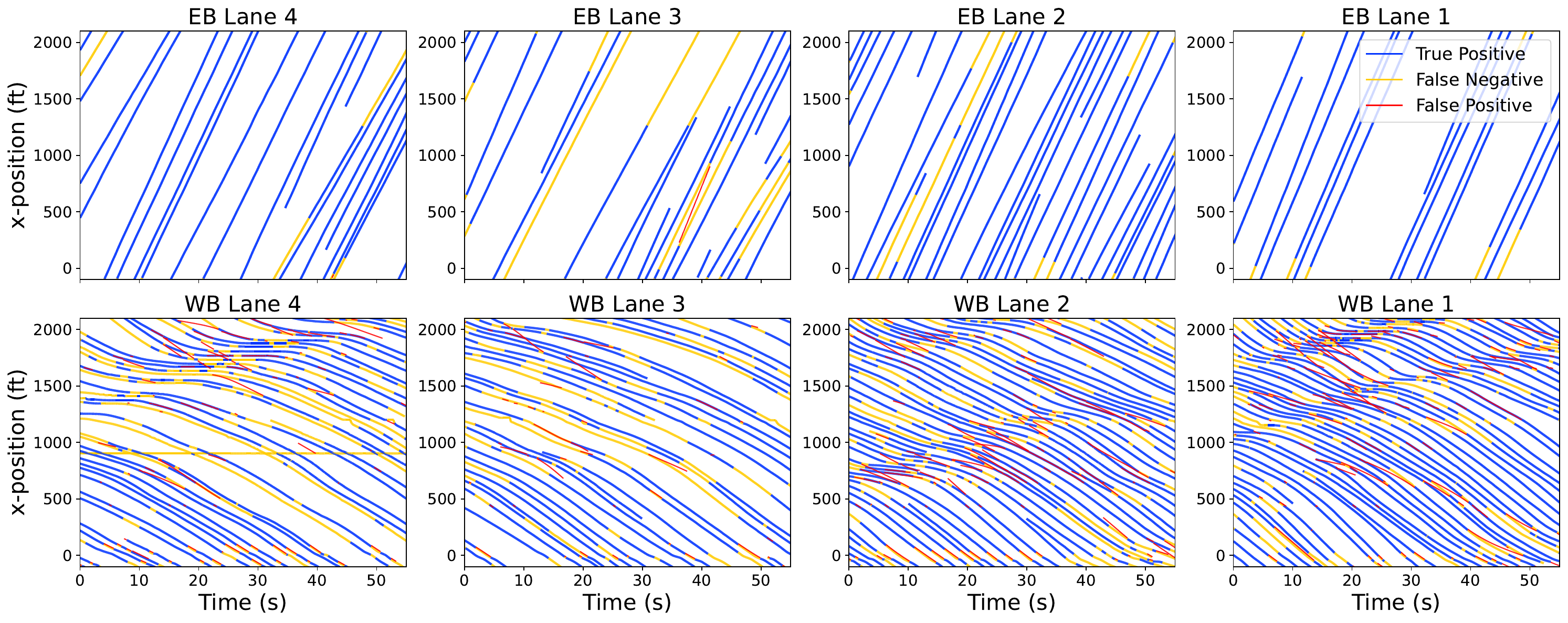}
    \caption{Time-space diagrams (object x-position vs time) for each lane for Dual3D +KIOU+TF pipeline on Scene 3 (Lane 4 is rightmost lane in direction of travel). False negatives (yellow), false positives (red) and true positives (blue) shown. In this case, most false positives are closely paired with a false negative, indicating that an object was tracked below the IOU threshold. Lanes farthest from cameras (EB lane 1 and WB lane 4) have the more false negatives in general, likely due to smaller object size and greater object occlusion. In some cases, a predicted object that falls below the IOU threshold with a ground truth object results in a parallel false positive and false negative track.}
    \label{fig:trajectories}
    \vspace{-0.2in}
\end{figure}

\section{Conclusion} 
\label{sec:conclusion}

This work introduced the I24-3D dataset, a multi-camera 3D vehicle tracking dataset with a total of 57 minutes of video and 877,000 vehicle annotations across 16-17 cameras. It also provided an initial benchmarking of some multi-camera 3D tracking pipelines from existing algorithms, demonstrating the difficulty of tracking on this dataset.

The benchmarking performed in this work represents a first step towards developing and evaluating efficient and accurate 3D multi-camera tracking pipelines. Moreover, though none of the benchmarked pipelines achieved performance suitable for fine-grained traffic analysis (i.e. HOTA > 0.75, mostly tracked objects > 95\%), we suspect that there do exist methods or combinations of methods that will perform better than the implemented methods from this work, especially those that better utilize the 3D scene information stemming from multiple cameras in an intensely occlusion-aware manner. We encourage interested researchers to report their results on this benchmark utilizing the protocol described in Section \ref{sec:experiments} and Appendix V. In the future, we look forward to developing such scene-aware MOT methods, armed with a new enabling dataset. We also intend to release a 3D multi-camera tracking challenge with new scenes and cameras from the I-24 MOTION system \cite{gloudemans202324}.

\bibliography{sources}
\end{document}

\maketitle

% \begin{abstract}
% This work presents a novel video dataset recorded from overlapping highway traffic cameras along an urban interstate, enabling multi-camera 3D object tracking in a traffic monitoring context. Data is released from 3 scenes containing video from at least 16 cameras, totaling 57 minutes in length. 877,000 3D bounding boxes and corresponding object tracklets are fully and accurately annotated for each camera field of view and are combined into a spatially and temporally continuous set of vehicle trajectories for each scene. Lastly, existing algorithms are combined to benchmark a number of 3D multi-camera tracking pipelines on the dataset, with results indicating that the dataset is challenging due to the difficulty of matching objects travelling at high speeds across cameras and  heavy object occlusion, potentially for hundreds of frames, during congested traffic. This work aims to enable the development of accurate and automatic vehicle trajectory extraction algorithms, which will play a vital role in understanding impacts of autonomous vehicle technologies on the safety and efficiency of traffic.
% \end{abstract}

 % A thorough error analysis is performed on the data, and sources of error arising from field deployments of IP cameras are corrected to improve the quality of the data.

%-------------------------------------------------------------------------
The following appendices are included in this supplement:
\begin{myitem}
\item \textbf{Appendix I:} File Format and Included Files
\item\textbf{Appendix II:} Scene Homography
\item \textbf{Appendix III:} Scene and Timestamp Errors and Corrections
\item \textbf{Appendix IV:} Vehicle Size Estimation Error Data
\item \textbf{Appendix V:} Additional Experimental Settings and Implementation Details
\item \textbf{Appendix VI:} Full Results
\item \textbf{Appendix VII:} Privacy Considerations
\end{myitem}

Additionally, the following supplementary files are included in this submission:
\begin{myitem}

\item \textbf{scene\textunderscore 1\textunderscore labels.csv} - annotations for Scene 1 provided as a file format example
\item \textbf{Scene1.mp4} - summary video for Scene 1 with all camera fields of view and object annotations drawn. Irrelevant portions of each frame are blurred. View at: \url{https://youtu.be/QKUL9Ify_OQ}
\item \textbf{Scene2.mp4} - summary video for Scene 2 with all camera fields of view and object annotations drawn. Irrelevant portions of each frame are blurred. View at: \url{https://youtu.be/t6bbNbwbkoY}
\item \textbf{Scene3.mp4} - summary video for Scene 3 with all camera fields of view and object annotations drawn. Irrelevant portions of each frame are blurred. View at: \url{https://youtu.be/UoPN6s07hbA}
\item \textbf{Scene3\textunderscore p3c5.mp4} - example video for a single camera from Scene 3 with object annotations drawn. Irrelevant portions of the frame are blurred. View at: \url{https://youtu.be/OcBsIzjuqNg}
\end{myitem}

\section*{Appendix I: File Format and Included Files}
The following files are included in the dataset for each scene:

\begin{myitem}
    \item \textbf{$\{$scene\textunderscore id$\}$\textunderscore labels.csv} - monocular 3D bounding boxes and vehicle class data for each vehicle in the area of interest, indexed by frame. csv columns are: \textit{frame index}, \textit{timestamp}, \textit{vehicle ID}, \textit{vehicle class}, \textit{vehicle x-position} (ft), \textit{vehicle y-position} (ft), \textit{vehicle length} (ft), \textit{vehicle width} (ft), \textit{vehicle height} (ft), \textit{direction of travel} (EB or WB), \textit{camera}.
    
    \item \textbf{$\{$scene\textunderscore id$\}$\textunderscore resampled.csv} - monocular 3D bounding boxes and vehicle class data for each vehicle in the area of interest, indexed by frame, and sampled at 30 Hz intervals from the best-fit spline for each trajectory to create a single continuous trajectory with continuous motion in the direction of travel. csv columns are: \textit{frame index}, \textit{timestamp}, \textit{vehicle ID}, \textit{vehicle class}, \textit{vehicle x-position} (ft), \textit{vehicle y-position} (ft), \textit{vehicle length} (ft), \textit{vehicle width} (ft), \textit{vehicle height} (ft), \textit{direction of travel} (EB or WB), \textit{camera}, and 16 values corresponding to the x and y pixel coordinates for each 3D bounding box corner coordinate (assuming 4K frame size).
   % Splines are of from the SciPy package $scipy.interpolate.UnivariateSpline$ \cite{2020SciPy-NMeth}.
    \item \textbf{$\{$scene\textunderscore id$\}$\textunderscore timestamps.csv} - original and corrected timestamps reported for each camera sequence, indexed by frame number in video sequence. Timestamps are reported in seconds since the last epoch and are quantized to 0.01s. csv columns are: \textit{frame index}, \textit{camera}, \textit{timestamp}, \textit{corrected timestamp}. 
    \item \textbf{$\{$scene\textunderscore id$\}$\textunderscore transforms/}
        \begin{itemize}
            \item \textbf{$\{$camera\textunderscore id$\}$\textunderscore $\{$direction$\}$\textunderscore homography.csv} - $\mathcal{P}$ and $\mathcal{H}$ parameters for image to roadway and roadway to image projection. The file consists of two lines of parameters as described in Appendix II: 
                            [$h_{11}$,$h_{12}$,$h_{13}$,$h_{21}$,$h_{22}$,$h_{23}$,$h_{31}$,$h_{32}$,$h_{33}$] 
                            [$p_{11}$,$p_{12}$,$p_{13}$,$p_{14}$,$p_{21}$,$p_{22}$,$p_{23}$,$p_{24}$,$p_{31}$,$p_{32}$,$p_{33}$,$p_{34}$]
                      
            \item \textbf{$\{$camera\textunderscore id$\}$\textunderscore $\{$direction$\}$\textunderscore curve.csv} - polynomial coefficients fitting curve used to offset y-coordinates according to roadway curvature, defined per direction of roadway travel. The three polynomial coefficients are listed in the first three columns of the first row. 
            \item \textbf{$\{$camera\textunderscore id$\}$\textunderscore $\{$direction$\}$.png} - image containing plotted correspondence points for homography, per direction of roadway travel
            \item \textbf{$\{$camera\textunderscore id$\}$\textunderscore mask.png} - a binary mask image with 1 for relevant portions of the image and 0 for other portions. This can be used to blur or mask irrelevant portions of the image during training/inference of models.

        \end{itemize}
    \item \textbf{$\{$scene\textunderscore id$\}$\textunderscore sequences/} 
        \begin{itemize}
            \item \textbf{$\{$camera\textunderscore id$\}$.mp4} -  4K resolution (3840$\times$2160) H.264-encoded video file recorded at 30 frames per second (nominal).
        \end{itemize}
\end{myitem}

% \section*{Appendix II: Links to Videos}
% A summary video for each sequence can be viewed at the links listed below. Each summary video contains all camera fields of view for the scene. An additional video showing 4K video for a single camera field of view is also included. All videos are trimmed to include 300 more frames than the (reported) number of annotated frames for the scene. 

% \Derek{LINKS!!!}
% \begin{myitem}
%     \item \textbf{Scene 1} - 
%     \item \textbf{Scene 2} - 
%     \item \textbf{Scene 3} - 
%     \item \textbf{Scene 3, p1c1} - 
% \end{myitem}

\section*{Appendix II: Scene Homography}
\subsection*{3D Perspective Transform Fitting}
A \textit{homography} relates two views of a planar surface. For each camera in each scene, we provide homography information such that the 8-corner coordinates of the stored 3D bounding-box annotation can be projected into any camera view for which the vehicle is visible, creating a monocular 3D bounding box within that camera field of view.  For each direction of travel in each camera view, for each scene, a homography relating the image pixel coordinates to the roadway coordinate system is defined. (Though the same cameras are used for different scenes, the positions of the cameras changes slightly over time due). A local flat plane assumption is used (the roadway is assumed to be piece-wise flat) \cite{hartley2003multiple}. A series of correspondence points series of correspondence points $p_q = [x,y,x',y',z']$ are used to define this relation, where $(x,y)$ is the coordinate of selected correspondence point $q$ in pixel coordinates (row, column) and $(x',y',z')$ is the selected correspondence point in roadway coordinates. 

All selected points are assumed to lie on the road plane, so $z' = 0$ for all selected correspondence points. Visible lane marking lines are used as correspondence points in each camera field of view. Each lane is reliably known to be 12 feet wide, and each lane-separating tick mark is known to be 10 feet long and at a regular spacing of 40 feet. Thus, the road plane coordinates of each lane tick mark are known precisely. The corresponding pixel coordinates are manually selected in each camera field of view, for each direction of travel on the roadway.

A \textit{perspective transform} (Equation \ref{eq:H}) is fit to these correspondence points. We first define a 2D perspective transform which defines a linear mapping (Equation \ref{eq:perspective}) of points from one plane to another that preserves straight lines. The correspondence points are then used to solve for the best perspective transform $\mathcal{H}$ as defined in equation \ref{eq:H}, where $s_i$ is a scale factor.

\begin{equation}
\label{eq:perspective}
    s_q \begin{bmatrix}
                x_q' \\
                y_q' \\
                1
        \end{bmatrix} 
        \sim
        \mathcal{H} 
            \begin{bmatrix}
                x_q \\
                y_q \\
                1
            \end{bmatrix}
\end{equation}

where $\mathcal{H}$ is a $3\times 3$ matrix of parameters:
\begin{equation}
\label{eq:H}
\mathcal{H} = 
            \begin{bmatrix}
            h_{11} & h_{12} & h_{13} \\
            h_{21} & h_{22} & h_{23} \\
            h_{31} & h_{32} & h_{33}
            \end{bmatrix} 
\end{equation}

 For each camera field of view and each direction of travel, the best perspective transform $\mathcal{H}$ is determined by minimizing the sum of squared re-projection errors according to equation \ref{eq:H_mse} as implemented in OpenCV's $find\mathunderscore homography()$ function~\cite{bradski2000opencv}:
\begin{equation}
\label{eq:H_mse}
   \min_\mathcal{H} \sum_q \left(x_q' - \frac{h_{11}x_q + h_{12}y_q + h_{13}} {h_{31}x_q + h_{32}y_q + h_{33}}\right)^2 + \left(y_q' - \frac{h_{21}x_q + h_{22}y_q + h_{23}} {h_{31}x_q + h_{32}y_q + h_{33}}\right)^2
\end{equation} 

The resulting matrix $\mathcal{H}$ allows any point lying on the plane within the camera field of view to be converted into roadway coordinates and, the corresponding matrix $\mathcal{H}_{inv}$ can easily be obtained to convert roadway coordinates on the plane into image coordinates.  However, since each vehicle is represented by a 3D bounding box, the top corner coordinates of the box do not lie on the ground plane. A 3D perspective transform $\mathcal{P}$ is needed to linearly map coordinates from 3D roadway space to 2D image coordinates, where $\mathcal{P}$ is a $3\times 4$ matrix of parameters:
\begin{equation}
\label{eq:P}
\mathcal{P} = 
            \begin{bmatrix}
            p_{11} & p_{12} & p_{13} & p_{14}\\
            p_{21} & p_{22} & p_{23} & p_{24}\\
            p_{31} & p_{32} & p_{33} & p_{34}
            \end{bmatrix} 
\end{equation}

and $\mathcal{P}$ projects a point in 3D space $(x',y',z')$ into the corresponding image point $(x,y)$ according to:

\begin{equation}
\label{eq:P_transform}
    \mathcal{P}  \begin{bmatrix}
                x' \\
                y' \\
                z' \\
                1
        \end{bmatrix} 
        \sim
            s'
            \begin{bmatrix}
                x \\
                y \\
                1
            \end{bmatrix}
\end{equation}

By observing the case where $z' = 0$, it is evident columns 1,2, and 4 of $\mathcal{P}$ are equivalent to the columns of $\mathcal{H}_{inv}$ and can be fit in the same way. Thus, we need only solve for column 3 of $\mathcal{P}$. Next, we note as in \cite{hartley2003multiple} that $(\frac{p_{11}}{p_{31}},\frac{p_{21}}{p_{31}})$ is the vanishing point (in image coordinates) of perspective lines drawn in the same direction as the roadway coordinate x-axis. The same is true for the 2nd column and the roadway coordinate y-axis, the 3rd column and the roadway coordinate z-axis, and the 4th column and the roadway coordinate origin.

Thus, to fully determine $\mathcal{P}$ it is sufficient to locate the vanishing point of the z-axis in roadway coordinates and to estimate the scaling parameter $p_{33}$. The vanishing point is located in image coordinates by finding the intersection point between lines drawn in the z-direction. Such lines are obtained by manually annotating vertical lines in each camera field of view. The scale parameter is estimated by minimizing the sum of squared re-projection errors defined in equation \ref{eq:P_mse} for a sufficiently large set of roadway coordinates and corresponding, manually annotated coordinates in image space.

\begin{align}
\label{eq:P_mse}
   \min_{p_{33}} \sum_q &\left(x_q - \frac{p_{11}x_q' + p_{12}y_q' + p_{13}z_q' + p_{14}} {p_{31}x_q' + p_{32}y_q' + p_{33}z_q' + p_{34}}\right)^2 + \nonumber \\ &\left(y_q - \frac{p_{21}x_q' + p_{22}y_q' + p_{23}z_q' + h_{24}} {p_{31}x_q' + p_{32}y_q' + p_{33}z_q' + h_{34}}\right)^2
\end{align} 
 
 The resulting 3D perspective transform $\mathcal{P}$ allows for the lossless conversion of points in roadway coordinates to the corresponding points in image coordinates.

\subsubsection*{Curvature Correction}

\begin{figure}[b]
    \centering
    \includegraphics[width = \textwidth]{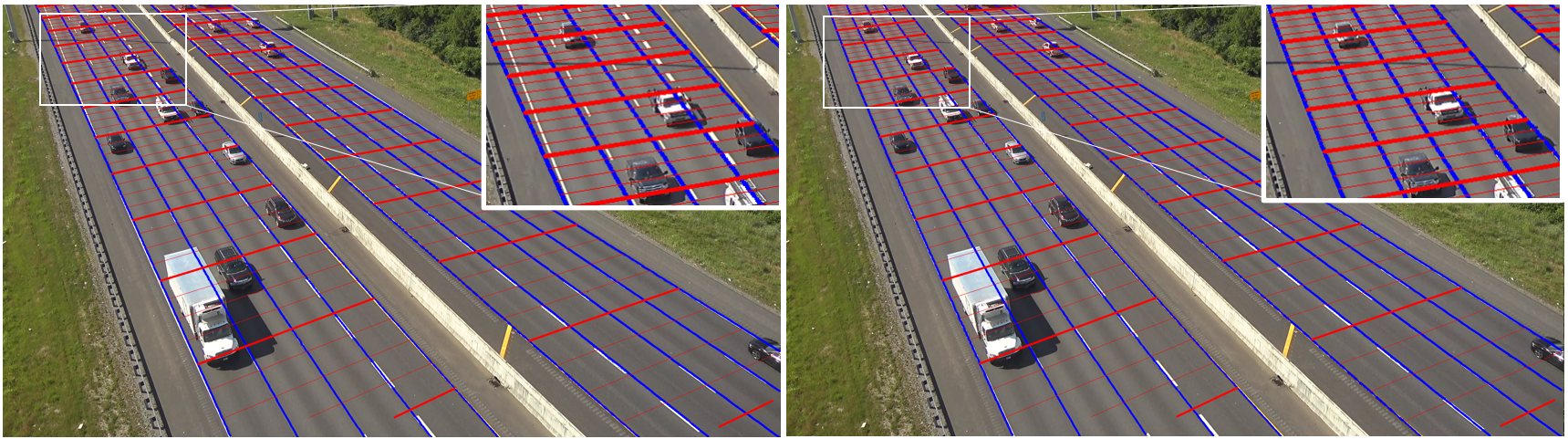}
    \caption{(Left) Before curvature correction, lines in the x-direction (blue) with equal y deviate significantly from roadway direction (white and yellow painted lines on roadway) at distances far from points used to fit local camera homography. (Right) After curvature correction, roadway lines in image have equal y-coordinate.}
    \label{fig:curve}
\end{figure}

Lastly, we fit a 2nd-order polynomial curve $f(x')$ to describe the y-coordinate along a solid lane line as a function of the x-coordinate (in space). We then use the fit curve to shift the y-coordinate of each point when it is transformed from image space to roadway coordinate space to account for lateral roadway curvature according to Equation \ref{eq:curve}.  

\begin{equation}
    \label{eq:curve}
    y_{offset}' = y' - f(x')
\end{equation}

where $y_{offset}'$ is the corrected y-coordinate and $f(x')$ is the fit 2nd-order polynomial. We analogously perform the reverse correction when converting points from 3D roadway coordinate space into 2D image space.  We note that the curvature also results in a slight error in x-coordinates, but the component of error in the x-direction is negligible for the purposes of this dataset as it is proportional to the sine of a relatively small angle (angle of roadway turn within the camera field of view). Figure \ref{fig:curve} shows coordinate system alignment to the roadway markings before and after curvature correction.

\section*{Appendix III: Sources of Timestamp Error and Corrections}
\textbf{We discover, through numerous tests, that the timestamps reported by the IP camera firmware, are inaccurate, approximately on the order of 0.1-1s.} This Appendix describes the types of error we identify, then details the corrections used to partially compensate for these errors. Lastly, it presents error metrics before and after these timestamp corrections.

\subsection*{Known Timestamp Errors}
\begin{figure}[H]
    \centering
    \includegraphics[width = \textwidth]{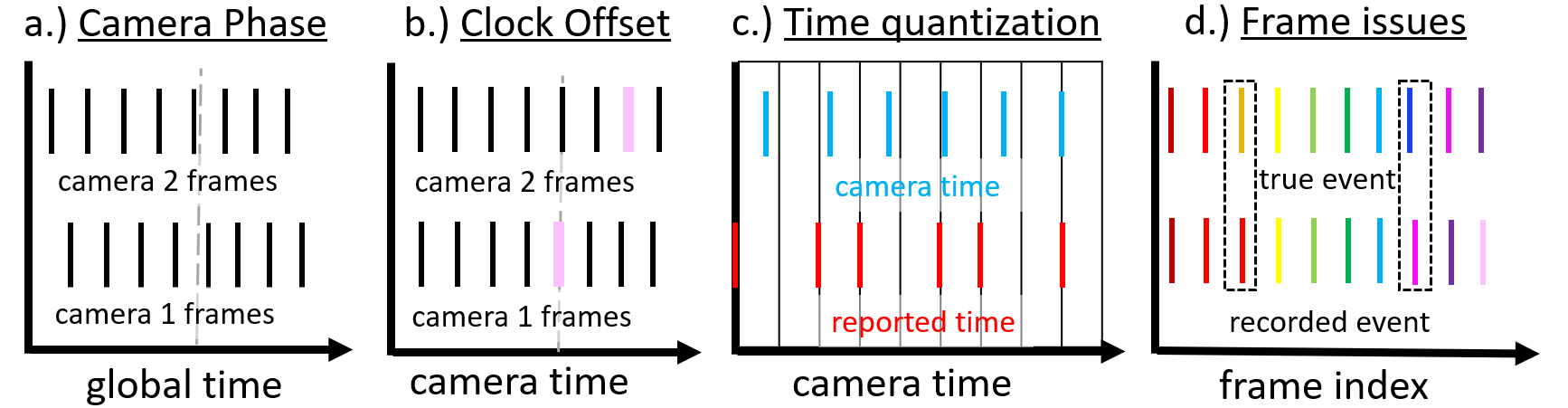}
    \caption{Multiple-camera timing issues. a.) Frames from different cameras are out of phase. b.) Different cameras report the same event (pink) as occurring at different times. c.) Camera time (blue) is quantized to a lower precision (red) before being reported. d.) Cameras report the same frame twice or skip frames (dash box).} 
    \label{fig:issues}
\end{figure}

Figure \ref{fig:issues} provides a visual overview of 4 timing errors (camera phase differences, camera clock offset, camera timestamp quantization, and doubled/skipped frames) in the I24-3D dataset. We then provide an example of each.

\begin{figure}[htb]
    \centering
    \includegraphics[width = \textwidth]{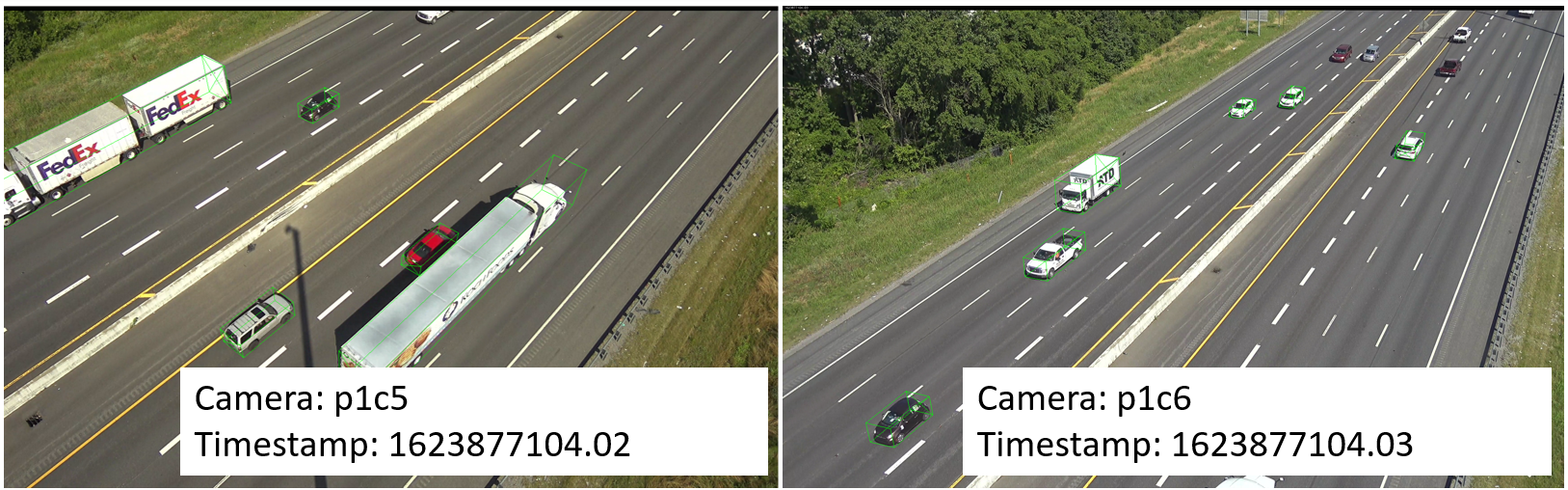}
    \caption{Example of phase and clock offset error. Closest reported timestamps from cameras are unequal (phase), and the positions of vehicles in camera p1c5 suggest this frame was recorded later than p1c5 (clock offset).}
    \label{fig:phase_and_offset}
\end{figure}

\textbf{Camera Phase and Offset:} Figure \ref{fig:phase_and_offset} shows the closest (in time) frames from camera p1c5 and p1c6; the reported timestamps for these frames are still 0.01s apart, indicating the cameras record frames out of phase. Furthermore, though camera p1c5 reports a time 0.01s earlier than p1c6, the position of vehicles suggest that this frame was recorded after the frame from p1c6 (vehicles are slightly further along the roadway in their respective directions of travel). We suspect this is due to camera clock offset (bias) relative to one another. 

\textbf{Camera Timestamp Quantization} Camera timestamps are reported to 0.01s precision, with higher-precision camera times quantized (rounded or truncated) before reporting. To demonstrate this behavior, we conduct the following test. We create two rotor clocks with two rotating disks (rotating at 2 Hz and 5 Hz). We control the rotational frequency of each rotor precisely using a dynamometer. We record the rotor clock rotation using a camera of the same model as deployed on I-24 MOTION (used to obtain the video data in this dataset). Based on the rotational position of each hand extracted using image processing techniques, we precisely determine the time \textit{deltas} ($\Delta$) (difference in time between consecutive frames).  We compare the time deltas  obtained by the coarse (2Hz) and fine (5Hz) rotor clocks to the reported timestamps from the camera. 

\begin{figure}[h]
    \centering
    \includegraphics[width = \textwidth]{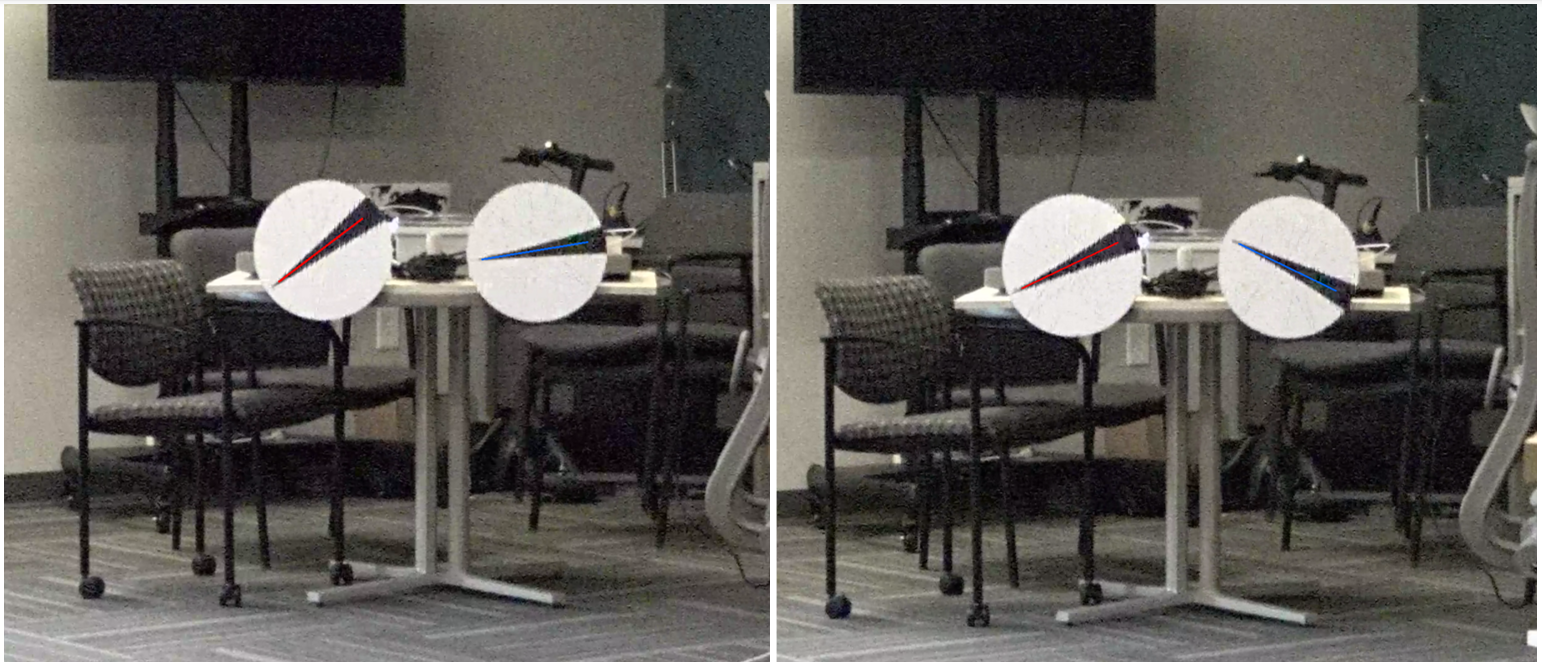}
    \caption{Coarse 2Hz (red) and fine 5Hz (blue) rotors shown for two frames of lab test video. The position of each rotor along with a total rotor revolution count can be used to determine the time since the start of the video sequence and the time delta between two consecutive frames.}
    \label{fig:clock}
\end{figure}

\begin{figure}[h]
    \centering
    \includegraphics[width = \textwidth]{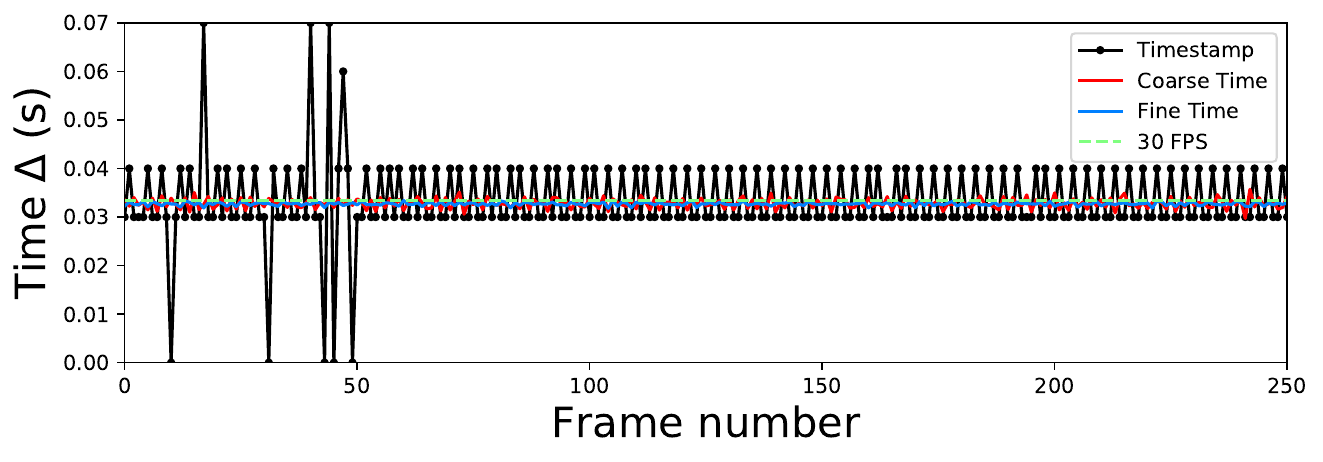}
    \caption{Time deltas between frames according to camera frame timestamps (black), coarse (red) and fine (blue) rotor clocks, compared to nominal framerate (green). Reported clock timestamps fluctuate between 0.03 and 0.04 s (with occasional skipped or doubled frames resulting in anomalous deltas). Rotor clock times show that the actual recording times for each frame adhere much more closely to nominal framerate.}
    \label{fig:deltas}
\end{figure}

Figure \ref{fig:clock} shows the clock setup with rotor positions, and Figure \ref{fig:deltas} shows the reported frame deltas. The true time at which frames are recorded can be estimated from the coarse and fine rotor clock times, which adhere closely to the 30 fps nominal framerate (subject to some jitter due either to slight camera deviations from the nominal framerate or slight errors in rotor positional extraction algorithm). Conversely, the reported clock quantized timestamp deltas fluctuate significantly around the true clock times. Doubled frames are visible (when delta = 0s, the same frame and timestamp has been sent twice by the camera). Thus, the quantization of timestamps is shown to create error in determining the true time at which a frame was recorded.

\begin{figure}[hb]
    \centering
    \includegraphics[width = \textwidth]{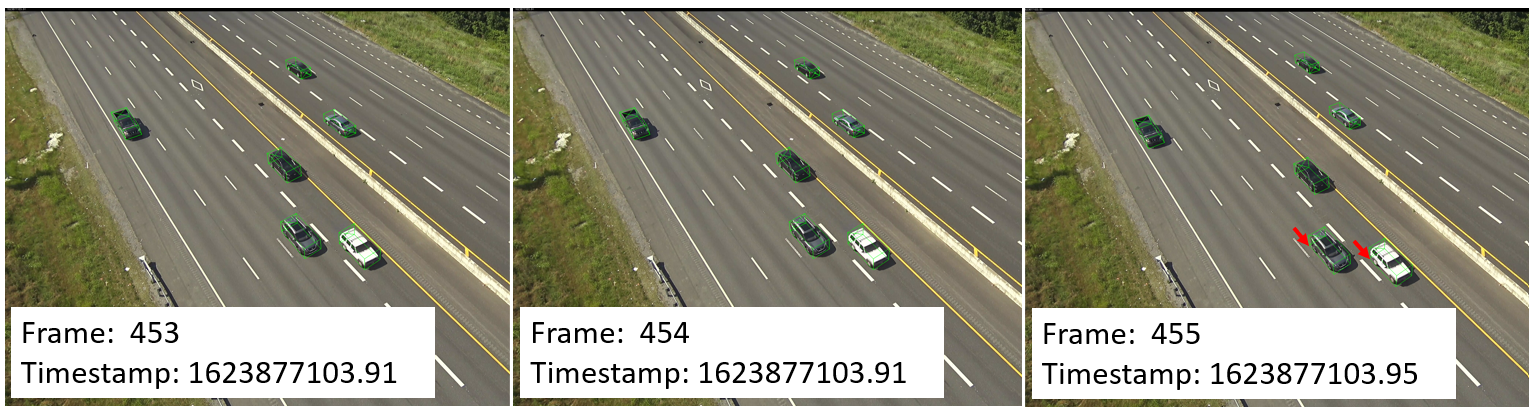}
    \caption{Example of a doubled frame. Frame indices 453 and 454 for camera p1c2 in sequence 0 display identical frames and timestamps. The following frames shows object positional changes (red arrows) consistent with a larger time delta than is indicated by frame timestamps.}
    \label{fig:double}
\end{figure}

\textbf{Skipped/Doubled Frames:} Lastly, Figure \ref{fig:double} shows an example of a doubled frame from camera p1c2. The same frame and timestamp are reported from the camera twice (frames 453 and 454). In frame 455, objects have moved 7-8 feet from their positions in frames 453 and 454, consistent with a time difference of roughly 2/30s (0.066s) for the average speeds at which the pictured vehicles are traveling, but the reported time delta is only 0.04s. This suggests that an inaccurate timestamp has been reported for frame 455.

\subsection*{Timestamp Error Correction}

To correct these timestamp errors, we perform 2 main operations and consider a third.

\subsubsection*{Time Offset Correction}
First we attempt to correct the camera clock offset or bias as we expect this value to be mostly fixed within a relatively short (1-2 min) time-frame. Let the estimate of the true time of frame $j$ for camera $k$ be denoted as $t_{j,k}'$, the reported camera timestamp be $t_{j,k}$the clock offset for camera $k$ be denoted as $o_k$, and the \textit{residual error} due to quantization and skipped/doubled frames for camera $k$ and frame $j$ denoted as $\epsilon_{jk}$. We estimate the true time as:
\begin{equation}
    t_{j,k}' = t_{j,k} + o_k + \epsilon_{j,k}
\end{equation}

Inspired by work on camera synchronization in sports \cite{kumar20113d}, we use vehicles viewed simultaneously by two cameras with overlapping fields of view to estimate the relative clock offset between the two ($o_k$ - $o_k-1$). Let $i$ index the set of $m$ objects  visible in a portion of the roadway coordinate system $[x_{i,min},x_{i,max]}$ in both cameras $k$ and $k-1$. We sample $s$ x-coordinates (indexed by $r$) uniformly spaced across the range $[x_{i,min},x_{i,max]}$. For each point in this range $x_r$, we linearly interpolate between the two closest annotations for camera $k$ to estimate the time at which camera $k$ would have reported object $i$ in position $x_r$. Let $\tau_{r,k}$ denote this time. We set $o_k - o_{k-1}$ to be the mean difference between estimated times of sample x-points from the two cameras.
\begin{equation}
    o_k  - o_{k-1} = \frac{\sum_i \sum_r (\tau_{r,k} - \tau_{r,k-1})}{m*s}
\end{equation}

and set $o_0$ = 0, allowing us to sequentially solve for the rest of camera clock errors. We ignore the effect of quantization for purposes of estimating the camera clock offsets. We assume that the per-camera clock offset does not drift over the relatively short duration of a single scene.

\subsubsection*{Residual Time Error Correction}
Next we address the remaining timestamp errors caused by quantization and skipped/double frames. To reason about out-of-phase camera frames, we utilize a continuous functional representation of each vehicle's trajectory. Inspired by the work of \cite{ren2018learning} and \cite{coifman2017critical}, we fit a cubic spline to each of the x position and the y position of vehicle annotations as a function of reported timestamp, which ensures that the resulting trajectory follows a constant jerk (3rd-derivative) model between spline \textit{knots} (points where polynomial coefficients of the spline function change). The number of knots $k_i$ for   is constrained based on reasonable driving assumptions according to equation \ref{eq:knots}:
\begin{equation}
\label{eq:knots}
    k_i \leq 2d_i
\end{equation}
where $d_i$ is the difference between the maximum and minimum timestamps for which an annotation for vehicle $i$ exists. 

The best-fit spline $f_{x,i}(t)$ for the x coordinates of object $i$ is found by minimizing the mean squared error between the spline-estimated object location and the annotated object location over all $j$ annotated boxes for object $i$ in all camera views:

\begin{equation}
    \min_{f_{x,i}} \sum_ j (w_{i,j}(f_{x,i}(t_j) - x_i,j))^2
\end{equation}

where $w_{i,j}$ is a per-annotation weighting factor equivalent to the shift in pixels resulting from a one foot change in the initial x-position of the annotation. This weighting enforces that the resulting spline distorts visual object positions within the frame as little as possible. A similar equation is used to determine the best fit spline for y coordinates $f_{y,i}(t)$ as a function of time.

We then use the best-fit spline $f_i^{(x)}(t)$ for x-position for each object $i$ visible within frame $j$ of camera $k$ to estimate the residual error for each camera frame $\epsilon_{j,k}$ according to Equation \ref{eq:quant}:

\begin{equation}
    \label{eq:quant}
    \min_{\epsilon_{j,k}} \sum_{i} (w_{i,j,k}(f_{i}^{(x)}(t_{j,k}) - x_{i,j,k}))^2
\end{equation}

where $w_{i,j,k}$ is the change (in pixels) resulting from a 1-foot change in x-position for object $i$ in frame $j$ of camera $k$. We constrain $\epsilon_{j,k}$ to be at most the change associated with a skipped frame ($1/30$s) plus quantization error (0.01s). 

\subsubsection*{Annotation shifting}
Thus far, adjustments have only altered timestamps. Further enhancements to smooth vehicle trajectories as they travel through multiple camera fields of view can be made by slightly altering the raw annotations at the expense of annotation accuracy within a frame. However, noting that the RMSE for human annotations is 6.24 pixels (see next section), we consider small adjustments to raw annotations within this range. We test allowable shifts of 1,2, and 3 pixels in the x and y direction separately such that each annotation is moved slightly towards the best-fit spline position at the corresponding time. For each allowable maximum shift, the average shift performed is much smaller (0.25px, 0.36px, 0.42px and 0.47px, respectively). Note that we ultimately chose not to use any annotation-shifting for the released dataset, prioritizing accuracy within each frame over cross-camera trajectory smoothness.

\subsection*{Post-Correction Error Characterization}
We characterize each known source of error in the dataset. When applicable, we characterize the error before and after timestamp corrections.

\textit{Human annotator variance} is estimated by labeling the same object multiple times for a selection of vehicles. The root mean-squared error (RMSE) is computed for all single-vehicle annotations, and this metric is averaged  across all sampled vehicles to estimate annotator root mean-squared error $RMSE_{ann}$ = 6.24 pixels for 4K resolution frames (3840$\times$2160 pixels). 

\textit{Vehicle size accuracy} is verified based on known vehicle (make and model) sizes. A subset (10\%) of labeled vehicles within the dataset are selected with discernible make and model. The annotated dimensions for each are compared to the actual size of that vehicle type when available, and otherwise to average vehicle size metrics for that class (semis and trucks). Over all sampled vehicles, a mean dimension error of -0.5ft, -0.1ft, and -0.2ft in length, width and height, respectively (standard deviation of 1.1ft, 0.3ft, and 0.6ft respectively). Mean errors for each dimension indicate only slight annotator bias towards undersized annotations, likely due to the ambiguous size of a curved 3D vehicle (vehicle classes with hard corners such as semis and trucks were not under-estimated). Full size comparison data is included in Appendix IV. 

\textit{Homography misalignment} is estimated by comparing the labeled positions of the same roadway point in multiple camera fields of view. Average across all camera fields of view, the average cross-camera projection error for homography matching points is 0.54ft/0.18ft (x/y directions) and 15.6 pixels for 4K video.

\textit{Cross-camera vehicle annotation misalignment} is computed by comparing vehicle annotations simultaneously visible in two or more cameras. We average the misalignment across all such annotations. Table \ref{tab:metric summary} reports the cross-camera vehicle annotation misalignment in the x ($CC_x$) and y ($CC_y$) directions (in feet) and the cross-camera pixel error ($CC_p$) after each correction. The corresponding errors for the points used to define the homography themselves are also displayed; it is unlikely that any correction could reduce error below that threshold without adjusting raw annotations since this error is added to the annotations during homography transformation.

\begin{table}
\centering
\begin{tabular}{lccc}
\toprule
\textbf{Correction} &\textbf{ $CC_x$} $\downarrow$ & \textbf{$CC_y$} $\downarrow$ &\textbf{ $CC_p$ }$\downarrow$  \\ \midrule
\textit{Homography} & 0.54 ft & 0.18 ft & 7.8 px  \\ \midrule
No Correction       & 1.52 ft & 0.69 ft & 35.2 px \\
Curve               & 1.52 ft & 0.44 ft & 29.6 px \\
Offset              & 1.39 ft & 0.44 ft & 22.2 px \\
Residual            & 1.24 ft & 0.44 ft & 15.6 px \\ \midrule
1px Shift          & 1.08 ft & 0.26 ft & 11.8 px \\
2px Shift          & 0.98 ft & 0.14 ft & 9.2 px \\
3px Shift          & 0.91 ft & 0.08 ft & 7.0 px \\ \bottomrule
\end{tabular}
\caption{All measured metrics for raw annotations and each sequential correction. Pixel errors are for boxes drawn on 4K-resolution frames (3840$\times$2160 pixels).}
\label{tab:metric summary}
\end{table}

\section*{Appendix IV: Vehicle Size Estimation Error Data}

Table \ref{tab:all_size_data} provides all data used to compute vehicle dimension accuracy statistics. Vehicle sizes were obtained from manufacturer websites when possible, or else estimated from class size averages (trucks and semis). Vehicle make, model and year estimated, with some mistakes in model and year possible due to difficulty in estimating this information from imagery. Vehicle size estimates were used from a 10\% subset of data selected across all dataset scenes.
{\setlength{\tabcolsep}{4pt}
\begin{longtable}{ccccccccc}
\toprule
\multicolumn{1}{l}{} &  &  & \multicolumn{3}{c}{Annotation (ft)} & \multicolumn{3}{c}{True (ft)} \\ \cmidrule(lr){4-6} \cmidrule(lr){7-9}
\multicolumn{1}{l}{ID} & Class & Vehicle Guess & \multicolumn{1}{c}{L} & \multicolumn{1}{c}{W} & \multicolumn{1}{c}{H} & \multicolumn{1}{c}{L} & \multicolumn{1}{c}{W} & \multicolumn{1}{c}{H} \\ \midrule
98 & midsize & 2018 Kia Soul & 13.3 & 5.6 & 5.4 & 13.6 & 5.9 & 5.3 \\
94 & sedan & 2014 Toyota Prius & 13.9 & 5.3 & 4.5 & 14.7 & 5.8 & 4.9 \\
0 & sedan & 2003 Chevrolet Impala & 15.9 & 5.7 & 4.5 & 16.7 & 6.1 & 4.8 \\
1 & pickup & 2017 Toyota Tacoma & 17.2 & 5.8 & 6.0 & 17.7 & 6.2 & 6.0 \\
64 & midsize & 2012 Toyota 4Runner & 16.1 & 6.0 & 5.8 & 15.8 & 6.3 & 6.0 \\
68 & midsize & 2018 Honda Odyssey & 16.4 & 6.5 & 5.9 & 16.9 & 6.6 & 5.8 \\
34 & semi & - & 73.8 & 9.0 & 13.3 & 72.0 & 8.5 & 13.5 \\
70 & semi & - & 76.1 & 8.6 & 14.8 & 72.0 & 8.5 & 13.5 \\
80 & midsize & 2018 Hyundai Tuscon & 14.6 & 6.2 & 6.1 & 14.7 & 6.1 & 5.4 \\
81 & pickup & 2015 Nissan Frontier & 17.5 & 6.3 & 5.7 & 17.2 & 6.1 & 5.8 \\
88 & midsize & 2004 GMC Yukon & 15.8 & 6.0 & 6.0 & 16.6 & 6.6 & 6.4 \\
96 & midsize & 2020 Toyota RAV4 & 15.3 & 5.9 & 6.0 & 15.2 & 6.1 & 5.8 \\
97 & midsize & 2020 Toyota RAV4 & 14.6 & 5.6 & 5.7 & 15.2 & 6.1 & 5.8 \\
83 & midsize & 2012 Honda CR-V & 14.4 & 5.9 & 4.9 & 14.8 & 6.0 & 5.4 \\
85 & pickup & 2016 Nissan Titan & 18.0 & 6.5 & 5.8 & 20.3 & 6.7 & 6.4 \\
40 & midsize & 2014 Chevrolet Equinox & 14.0 & 6.0 & 5.0 & 15.7 & 6.1 & 5.5 \\
41 & midsize & 2014 Toyota Highlander & 15.2 & 6.4 & 5.0 & 15.9 & 6.3 & 5.7 \\
39 & sedan & 2013 Honda Accord & 14.9 & 5.8 & 4.0 & 15.8 & 6.1 & 4.8 \\
34 & semi & - & 73.8 & 9.0 & 13.3 & 72.0 & 8.5 & 13.5 \\
32 & midsize & 2014 Dodge Caliber & 13.6 & 6.0 & 4.9 & 14.5 & 5.8 & 5.0 \\
28 & sedan & 2017 Honda Accord & 15.1 & 6.0 & 4.4 & 15.8 & 6.1 & 4.8 \\
17 & van & 2010 Chevrolet Express & 18.0 & 6.5 & 6.9 & 18.7 & 6.6 & 6.9 \\
15 & pickup & 2016 Ford F150 & 18.0 & 6.5 & 6.1 & 19.3 & 6.7 & 6.3 \\
57 & sedan & 2015 fiat 500 & 11.0 & 5.4 & 4.6 & 11.7 & 5.3 & 4.9 \\
58 & truck & 17 foot U-Haul style Box Truck & 23.3 & 8.4 & 9.6 & 23.9 & 7.7 & 10.0 \\
74 & midsize & 2015 Chrysler Town and Country & 15.5 & 6.3 & 5.0 & 16.9 & 6.6 & 5.7 \\
9 & truck & 20 foot U-Haul style Box Truck & 26.8 & 8.0 & 13.3 & 26.6 & 7.7 & 10.1 \\
87 & midsize & 2012 Honda CR-V & 14.4 & 5.7 & 4.9 & 14.8 & 6.0 & 5.4 \\
90 & sedan & 2014 Nissan Altima & 15.4 & 6.1 & 4.1 & 16.0 & 6.0 & 4.8 \\
63 & truck & 17 foot U-Haul style Box Truck & 25.8 & 8.3 & 11.9 & 23.9 & 7.7 & 10.0 \\
106 & sedan & 2004 Cadillac Deville & 16.6 & 5.7 & 4.3 & 17.3 & 6.3 & 4.8 \\
108 & van & 2012 Chevrolet Express & 19.2 & 6.5 & 6.3 & 18.7 & 6.6 & 6.8 \\
109 & midsize & 2012 Honda CR-V & 14.2 & 5.5 & 5.0 & 14.8 & 6.0 & 5.4 \\
113 & pickup & 2014 Dodge 1500 & 18.7 & 6.5 & 5.8 & 19.1 & 6.6 & 6.3 \\
95 & semi & - & 72.1 & 8.9 & 12.8 & 72.0 & 8.5 & 13.5 \\
26 & van & 2012 Chevrolet Express & 17.8 & 6.5 & 7.0 & 18.7 & 6.6 & 6.8 \\
27 & midsize & Jeep Grand Cherokee & 13.8 & 5.8 & 5.6 & 15.8 & 6.3 & 5.8 \\
1 & van & 2018 Ford Transit 250 & 20.6 & 6.3 & 9.9 & 22.2 & 6.8 & 9.1 \\
0 & pickup & 2018 Toyota Tacoma & 18.0 & 6.5 & 6.5 & 18.8 & 6.3 & 6.0 \\
7 & semi & - & 72.6 & 8.7 & 13.1 & 72.0 & 8.5 & 13.5 \\
73 & midsize & 2018 Nissan Rogue & 14.8 & 6.3 & 5.4 & 15.4 & 6.0 & 5.7 \\
13 & midsize & 2004 Honda CR-V & 13.0 & 5.4 & 6.3 & 14.9 & 5.8 & 5.5 \\
93 & midsize & 2018 Kia Soul & 12.4 & 6.0 & 5.0 & 14.3 & 5.8 & 5.3 \\
257 & van & 2007 GMC Savana & 19.2 & 6.3 & 6.8 & 18.7 & 6.6 & 6.8 \\
34 & midsize & 2015 Chevrolet Suburban & 17.2 & 6.7 & 5.5 & 18.5 & 6.6 & 6.4 \\
248 & pickup & 2018 Chevrolet Silverado 1500 & 19.2 & 6.4 & 6.0 & 20.0 & 6.7 & 6.2 \\
67 & sedan & 2016 Kia Forte & 14.0 & 5.8 & 4.5 & 15.0 & 5.8 & 4.7 \\
117 & midsize & 2014 Chevrolet Equinox & 14.4 & 5.9 & 5.2 & 15.7 & 6.1 & 5.5 \\
116 & semi & - & 73.5 & 8.2 & 13.7 & 72.0 & 8.5 & 13.5 \\
64 & sedan & 2012 Nissan Altima & 14.6 & 5.9 & 4.3 & 15.9 & 5.9 & 4.8 \\
32 & midsize & 2016 Jeep Wrangler (4 door) & 14.2 & 6.3 & 5.5 & 15.3 & 6.2 & 6.1 \\
114 & midsize & 2015 Dodge Grand Caravan & 16.0 & 6.0 & 5.5 & 16.9 & 6.6 & 5.8 \\
4 & sedan & 2014 Toyota Prius & 13.8 & 5.4 & 4.5 & 14.7 & 5.8 & 4.9 \\
90 & midsize & 2018 Honda Fit & 12.8 & 5.4 & 4.5 & 13.4 & 5.6 & 5.0 \\
0 & van & 2016 Dodge Sprinter & 18.2 & 6.4 & 7.5 & 19.4 & 6.7 & 7.8 \\
10 & truck & 17 foot U-Haul style Box Truck & 24.8 & 6.8 & 10.4 & 23.9 & 7.7 & 10.0 \\
245 & midsize & 2012 Chevrolet HHR & 13.6 & 5.4 & 5.0 & 14.7 & 5.8 & 5.3 \\
238 & van & 2015 Ford Transit & 17.2 & 6.3 & 7.1 & 18.3 & 6.8 & 7.0 \\
98 & sedan & 2017 Chrysler 300 & 15.3 & 6.2 & 4.2 & 16.6 & 6.3 & 4.9 \\
140 & truck & 17 foot U-Haul style Box Truck & 24.0 & 7.9 & 9.5 & 23.9 & 7.7 & 10.0 \\
183 & semi & - & 72.8 & 8.8 & 12.7 & 72.0 & 8.5 & 13.5 \\
101 & midsize & 2018 Honda Fit & 12.2 & 5.2 & 4.7 & 13.4 & 5.6 & 5.0 \\
182 & midsize & 2014 Chevrolet Equinox & 14.2 & 6.2 & 5.1 & 15.7 & 6.1 & 5.5 \\
95 & sedan & 2015 Chevrolet Malibu & 15.0 & 6.1 & 4.2 & 16.0 & 6.0 & 4.8 \\
137 & sedan & 2017 Ford Fiesta & 12.2 & 5.3 & 4.7 & 13.3 & 5.7 & 4.8 \\
94 & sedan & 2014 Toyota Corolla & 14.8 & 6.0 & 4.2 & 15.3 & 5.8 & 4.8 \\
92 & sedan & 2015 Chevrolet Cruze & 14.2 & 5.8 & 4.2 & 15.1 & 5.9 & 4.8 \\
176 & sedan & 2016 Nissan Leaf & 11.2 & 5.4 & 4.6 & 14.6 & 5.8 & 5.1 \\
134 & pickup & 2016 Chrevolet Silverado 1500 & 18.0 & 6.6 & 5.3 & 17.1 & 6.7 & 6.2 \\
214 & midsize & 2018 Jeep Compass & 13.4 & 5.7 & 5.0 & 14.4 & 6.2 & 5.4 \\
224 & midsize & 2020 Toyota RAV4 & 14.0 & 6.2 & 5.3 & 15.2 & 6.1 & 5.8 \\
229 & pickup & 2016 Ford F150 Crew Cab & 18.6 & 6.8 & 5.7 & 19.3 & 6.7 & 6.3 \\ \bottomrule
\caption{Annotated and true dimensions for assessed data subset (10\% of vehicles).}
\label{tab:all_size_data}
\end{longtable}

Table \ref{tab:size2} reports aggregate size estimate error metrics for each class and averaged over all samples. For each dimension (length, width and height) the percentage of vehicles that are reported with less than 1 foot of error in this dimension is summarized, as is the mean error and standard deviation in annotations errors).  Over all sampled vehicles, a mean dimension error of -0.5ft, -0.1ft, and -0.2ft in length, width and height, respectively, was obtained (annotations were too small on average). The standard deviation for each error was 1.1ft, 0.3ft, and 0.6ft respectively. Mean errors for each dimension indicate only slight annotator bias towards undersized annotations, likely due to the difficult of exactly sizing curved 3D vehicles from an oblique angle (notably, vehicles with hard corners such as semis and trucks were not under-estimated). The dimension with the largest distribution of error was length (1.1ft standard deviation).  Notably, 96\% of vehicle annotations have height accurate within 1 foot and 100\% of vehicle annotations have width accurate within 1 foot.

\begin{table}[ht]
\centering
\begin{tabular}{lcccc}
\toprule
Class & Samples & Length (ft) & Width (ft) & Height (ft) \\ \midrule
sedan   & 16                          & -1.0                 (0.7)                  & -0.2                 (0.2)                  & -0.5                 (0.2)                  \\
midsize & 28                          & -0.9                 (0.6)                  & -0.2                 (0.3)                  & -0.3                 (0.4)                  \\
van     & 7                           & -0.6                 (0.7)                  & -0.2                 (0.2)                  & +0.0                  (0.5)                  \\
pickup  & 9                           & -0.6                 (0.9)                  & -0.1                 (0.2)                  & -0.3                 (0.4)                  \\
truck   & 5                           & +0.5                  (0.9)                  & +0.2                  (0.6)                  & 0.9                  (1.6)                  \\
semi    & 7                           & +1.5                  (1.3)                  & +0.2                  (0.3)                  & -0.1                 (0.7)                  \\ \midrule

Under 1 ft Error & - & 60\% & 100\% & 96\% \\ \midrule

\textbf{Total}   & \textbf{72}        & \textbf{-0.5}\textbf{(1.1)}        & \textbf{-0.1}         \textbf{(0.3)}         & \textbf{-0.2 }       \textbf{(0.6) }    \\ \bottomrule
\end{tabular}
\caption{Annotation dimension error mean (standard deviation) compared to known vehicle sizes. + indicates mean estimated dimension is too large.}
\label{tab:size2}
\end{table}

\section*{Appendix V: Additional Experimental Settings and Implementation Details}

\subsection*{Evaluation Protocol}
To assess the difficulty of the tracking dataset and to provide initial evidence on the suitability of existing tracking algorithms, we benchmark a set of tracking methods on this dataset. Experimental protocol, metrics for evaluation, and implemented algorithms are described in this Appendix.

\subsubsection*{Model Training}
Each scene is split into temporally contiguous training and validation partitions (the first 80\% and the last 20\% of each scene, respectively). Thus, the validation partition of Scene 1 consists of all frames greater than or equal to 2160 from each sequence, and the validation partition consists of all frames greater than or equal to 1440 for Scenes 2 and 3. Detection model training is performed exclusively using the training partition. All training is performed locally on RTX6000 GPUs, and detection models are trained until convergence (Generally 10-15 epochs). 

\subsubsection*{Tracking}
Camera frames are out of phase (see Appendix III). To account for this, during tracking we maintain tight 1/60th second synchronization between videos during tracking using corrected frame timestamps, skipping frames as necessary to nominally maintain a 15 Hz frame rate (empirically, performance degrades above this frame rate due to increased false positives). We perform tracking across all video sequences for a scene in parallel (that is, all detections and tracklet updates are performed across all cameras for the same approximate time). This is required for crop-based tracking pipelines \cite{gloudemans2021vehicle} but not for tracking-by-detection pipelines. Tracking is performed through the entire scene duration (i.e. across the training and validation partitions of the dataset). 

\subsubsection*{Detector AP Testing}
We evaluate each detector on the validation partition of the I24-3D dataset. We follow the procedure detailed in The Pascal VOC Dataset Challenge guidelines \cite{everingham2009pascal}, including penalizing for multiple predicted objects corresponding to the same ground truth object. To filter objects, we remove all objects with output confidence less than 0.01, and perform non-maximal suppression on the set of detection outputs, first in the image coordinate system with a requisite IOU of 0.4 for removal, then in roadway coordinates with a requisite IOU of 0.1 for removal. We compare the remaining set of detections against the ground truth detections in roadway coordinates. We use \textit{birds-eye view precision} ($AP_{bev}$) rather than \textit{3D precision} ($AP_{3D}$) (i.e. height is not included in the evaluation). The results of this test are shown in Appendix VI.

\subsubsection*{Tracking Evaluation}
We evaluate each tracking pipeline on each entire scene, including both the training and validation partitions. We fit a best-fit 3rd order polynomial spline to each ground truth object to obtain a continuous object representation in roadway coordinates as described in Appendix III. Predicted vehicle trajectories are compared against boxes sampled from the best-fit spline for each object. Since ground truth objects are labeled in cameras with varying start times and tracked objects are produced with synchronized timestamps, we compute and compare only the temporally-overlapping sections of each ground truth trajectory dataset with the predictions. We linearly interpolate between the spline-sampled boxes and the tracker-output predictions at 30Hz to produce object sets at the same discrete times.

\textbf{IOU Threshold:} For all metrics except HOTA (which uses a variable IOU threshold for considered matches), we use a required IOU of 0.3 throughout evaluation. This is because, despite our best efforts to fully rectify annotations corresponding to the same vehicle viewed from different cameras at the same time, we are not able to fully remove the projection errors between these annotations (see Appendix III). Thus, we seek to avoid penalizing tracking algorithms for output errors that could reasonably be an artifact of these inconsistencies, so we select a somewhat lax IOU threshold to account for 1.24ft/0.44ft X/Y cross-camera annotation misalignment. 

\subsection*{Algorithm Implementation Details}
\subsubsection{3D Detectors}
\begin{myitem}
    \item \textbf{Monocular 3D Detector (Single3D)} - a Retinanet model with Resnet34-FPN backbone \cite{lin2017focal}. The outputs from this network are parameterized as a rectangular prism rather than as corner coordinates, which empirically leads to better model convergence. The formulation is camera-agnostic (as training a separate model for each camera FOV is infeasible both from data scarcity and scalability standpoints.) We remove all detections with a confidence lower than 0.3, and perform non-maximal suppression on detection outputs per camera, in pixel coordinates, with an IOU threshold of 0.4, then in shared roadway coordinates with an IOU threshold of 0.01. We quantize the model to half precision (16-bit float) for speed at inference. Code is originally from \url{https://github.com/yhenon/pytorch-retinanet}.
    \item \textbf{Monocular 3D Multi-frame Detector (Dual3D)} - Inspired by recent works utilizing multiple frames for detection and tracking \cite{zhou2020tracking}, we add the previous frame as detection input. We double the input channels of the model's first convolutional layer to accomodate the additional input. As above, we remove all detections with a confidence lower than 0.3, and perform non-maximal suppression on detection outputs per camera, in pixel coordinates, with an IOU threshold of 0.4.  We quantize the model to half precision (16-bit float) for speed at inference.
    \item \textbf{Monocular 3D Crop Detector (CBT)} - as described in \cite{gloudemans2021vehicle}, we train a Retinanet Model with Resnet34-FPN backbone for detecting objects in cropped portions of full frames. We expand each object prior by 1.3x to select the relevant pixels of a frame for each object, and resize these pixels to a crop size of 112x 112 pixels. After detection, we keep only the 50 highest confidence outputs from the detector and then using a weighting factor $W$ of 0.4 to weight confidence and IOU with object prior to select the best detection for each existing object. On full-frame detections frames (every 4 frames), we use the Dual3D detector.
    \item \textbf{Ground Truth Detections (GT)} - perfect ground-truth detections, stored natively in roadway coordinates.
\end{myitem}
\vspace{-0.2in}

\subsubsection{Object Trackers}
\begin{myitem}
    \item \textbf{Kalman-Filter IOU Tracker (KIOU)} - as described in \cite{bochinski2017high}. We utilize a contant velocity roadway-coordinate Kalman filter for object position prediction. We use the object-to-detection intersection over union metric in roadway coordinates to select the best-matching detection for each existing object.
    \item \textbf{ByteTracker (Byte)} - we utilize the two-stage association method described in \cite{zhang2022bytetrack}, using IOU as both primary and secondary matching criterion and utilizing a Kalman filter as suggested by authors. As suggested by the authors, we relax the criteria of each detector such that all objects with confidence higher than 0.3 are kept for the primary matching step and all object with confidence between 0.01 and 0.3 are kept for the secondary matching phase.
    \item \textbf{Crop-based Tracking (CBT)} - as proposed in \cite{gloudemans2021vehicle}, detection on some frames is performed by re-detecting priors in cropped subsets of the overall frame, and object associations are implicit for these frames. 
    \item \textbf{Ground Truth Single Camera Tracklets} - perfect single-camera tracklets.
\end{myitem}
\vspace{-0.15in}

\subsubsection{Cross-Camera Rectification Method}
\begin{myitem}
    \item \textbf{Detection Fusion (DF)} - as preferred in the AV context \cite{caesar2020nuscenes}, detections from all cameras are combined online in roadway coordinates and non-maximal suppression with a stringent 0.01 IOU threshold utilized to eliminate overlapping detections. 
    \item \textbf{Trajectory Stitching (TF)} - as proposed in \cite{wang2022automatic}, single camera tracklets are compared for spatio-temporal overlap offline, stitched together when a matching criteria is met, and refined to optimally describe the observed set of tracked object positions. We refer the interested reader to the cited work for an explanation of parameters and their meanings. We use this algorithm as implemented at \url{https://github.com/yanb514/I24-postprocessing}.
    \item \textbf{None} - as a baseline, object tracklets from each camera are output with no fusion.
    \item \textbf{Both (DF+TF)} - Tracking uses detection fusion, and a subsequent trajectory stitching step is performed to deal with remaining object fragmentations.
\end{myitem}

\section*{Appendix VI: Full Results}
This Appendix details AP testing results, tracking results for  pipelines utilizing ground truth detections or tracklets as input, and finally lists per-scene results for all pipelines.

\subsection*{Detector AP Testing Results}

Figure \ref{fig:AP} show the results of detector average precision testing. At all tested IOU thresholds, the Dual3D network has the highest AP score (0.572 $AP_{70}$), and the Single3D model has the lowest AP score at all 3 thresholds (0.254 $AP_{70}$). Interestingly, despite the large difference in detection precision, comparable tracking pipelines using these two detectors show only slight or negligible performance difference, perhaps explainable by insufficient Kalman filter parameter tuning to account for the more accurate measurements provided by the Dual3D detector. 

\begin{figure}[H]
    \centering
    \includegraphics[width=0.9\textwidth]{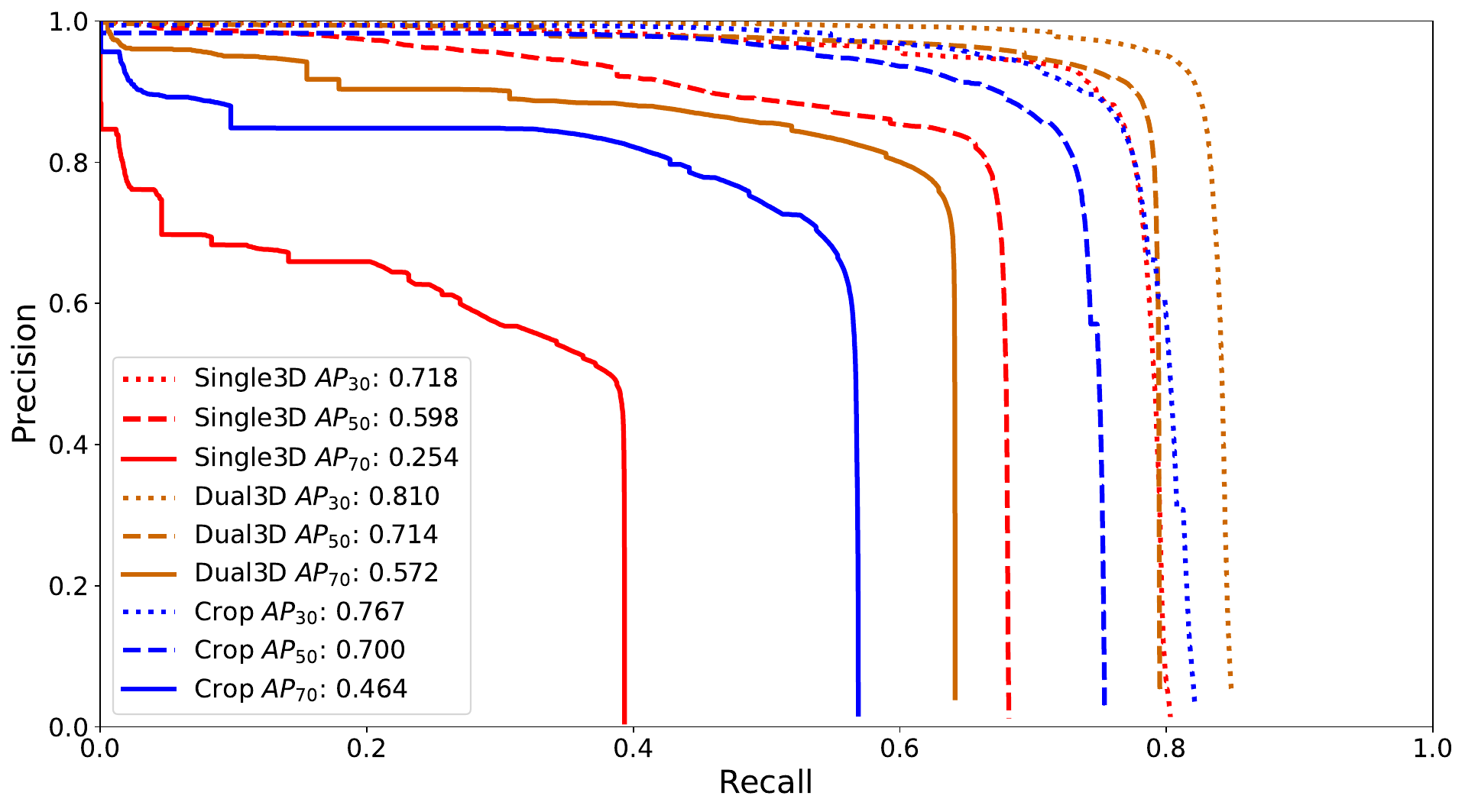}
    \caption{Precision versus recall curves for each detector at IOU thresholds of 0.7 (solid), 0.5 (dash) and 0.3 (dot), generated as in \cite{everingham2009pascal}. Overall AP score for each model at each threshold is listed in the legend.}
    \label{fig:AP}
\end{figure}

\subsection*{ Results for Ground Truth Pipelines}
Table \ref{tab:GT-detections} shows per-scene and average results for each implemented pipeline using ground truth detections as input. A few results are of note. Trajectory Fusion (TF) is the most accurate multi-camera rectification method, with Detection Fusion (DF) and DF + TF slightly less accurate and roughly equal across both trackers (KIOU and ByteTrack). Though recall is high (89\%) and precision is very high (as high as 99.8 \%), still only 73.6\% of objects are fully tracked at best. These results show that high-quality detections alone are not enough to solve the multi-camera tracking problem.

Table \ref{tab:GT-tracklets} shows results for all pipelines utilizing ground truth single-camera tracklets, either as-is or with a subsequent trajectory fusion (TF) step. With trajectory fusion, an HOTA of 61.7\% is achieved. This score is mostly driven down by remaining ID switches (0.51 per ground truth object on average), as the localization accuracy (MOTP) is fairly high (83.6\% average). Intuitively, it makes sense that given great single-camera tracklets, the remaining difficulties are caused by ID switches across cameras which constitute failures in trajectory fusion. The results of the AP testing, ground truth detection and tracklet pipelines indicate that there is room for improvement in all 3 components of the implemented multi-camera pipelines (detector, tracker and multi-camera rectification).

\begin{table}[H]
\setlength{\tabcolsep}{1.5pt} % Default value: 6pt
\renewcommand{\arraystretch}{0.8}

\centering
\begin{tabular}{lcc|l|cccccccccc} \toprule
\textbf{Tra.} & \textbf{DF} & \textbf{TF} & \textbf{Scene} & \textbf{HOTA} & \textbf{MOTA} & \textbf{MOTP} & \textbf{Rec} & \textbf{Prec} & \textbf{GT\%} & \textbf{Pred\%} & \textbf{MT} & \textbf{ML} & \textbf{Sw/GT} \\ \toprule
Byte & \checkmark & \checkmark & 1 & 66.5 & 94.0 & 78.5 & 94.3 & 99.7 & 96.6 & 100.0 & 89.8 & 0.9 & 0.13 \\
Byte & \checkmark & \checkmark & 2 & 47.3 & 85.6 & 62.2 & 88.0 & 97.4 & 86.0 & 100.0 & 69.3 & 3.5 & 0.23 \\
Byte & \checkmark & \checkmark & 3 & 40.9 & 76.3 & 82.6 & 79.1 & 96.8 & 92.5 & 99.8 & 57.7 & 5.3 & 1.30 \\
Byte & \checkmark & \checkmark & \textit{avg} & 51.6 & 85.3 & 74.4 & 87.1 & 98.0 & 91.7 & 99.9 & 72.2 & 3.3 & 0.55 \\ \midrule 

KIOU & \checkmark & \checkmark & 1 & 66.3 & 92.2 & 78.5 & 93.2 & 99.0 & 96.0 & 100.0 & 89.8 & 2.2 & 0.11 \\
KIOU & \checkmark & \checkmark & 2 & 48.2 & 87.0 & 62.0 & 89.0 & 97.9 & 86.0 & 100.0 & 73.7 & 4.4 & 0.20 \\
KIOU & \checkmark & \checkmark & 3 & 38.3 & 65.7 & 69.9 & 74.8 & 89.2 & 82.6 & 93.0 & 44.5 & 19.9 & 0.59 \\
KIOU & \checkmark & \checkmark & \textit{avg} & 50.9 & 81.7 & 70.1 & 85.7 & 95.4 & 88.2 & 97.7 & 69.3 & 8.8 & 0.30 \\ \midrule

Byte &  & \checkmark & 1 & 71.7 & 95.5 & 78.5 & 95.5 & 99.9 & 96.6 & 100.0 & 91.9 & 0.9 & 0.00 \\
Byte &  & \checkmark & 2 & 69.2 & 89.3 & 86.3 & 90.4 & 98.8 & 86.0 & 100.0 & 68.4 & 3.5 & 0.22 \\
Byte &  & \checkmark & 3 & 37.8 & 73.1 & 82.3 & 80.8 & 91.5 & 93.6 & 95.8 & 60.1 & 4.3 & 1.64 \\
Byte &  & \checkmark & \textit{avg} & \textbf{59.6 }& \textbf{86.0} & 82.4 & 88.9 & 96.8 & \textbf{92.0} & 98.6 & 73.5 & \textbf{2.9 }& 0.62 \\ \midrule 

KIOU &  & \checkmark & 1 & 71.7 & 95.5 & 78.5 & 95.5 & 99.9 & 96.6 & 100.0 & 91.9 & 0.9 & 0.00 \\
KIOU &  & \checkmark & 2 & 68.7 & 89.1 & 86.3 & 90.2 & 98.9 & 86.0 & 100.0 & 67.5 & 3.5 & 0.23 \\
KIOU &  & \checkmark & 3 & 38.1 & 73.4 & 82.3 & 81.4 & 91.3 & 93.6 & 95.4 & 61.2 & 4.3 & 1.63 \\
KIOU &  & \checkmark & \textit{avg} & 59.5 & \textbf{86.0} & 82.4 & \textbf{89.0} & 96.7 & \textbf{ 92.0} & 98.5 & \textbf{ 73.6} & \textbf{2.9} & 0.62 \\ \midrule
 
Byte & \checkmark &  & 1 & 63.0 & 91.3 & 90.9 & 91.5 & 100.0 & 96.6 & 100.0 & 85.4 & 0.9 & 0.69 \\
Byte & \checkmark &  & 2 & 52.4 & 82.8 & 92.0 & 83.4 & 99.5 & 86.0 & 100.0 & 58.8 & 3.5 & 1.02 \\
Byte & \checkmark &  & 3 & 38.8 & 76.7 & 91.6 & 77.0 & 100.0 & 93.2 & 100.0 & 52.3 & 5.3 & 2.17 \\
Byte & \checkmark &  & \textit{avg} & 51.4 & 83.6 & 91.5 & 84.0 & \textbf{99.8} & 91.9 & \textbf{100.0} & 65.5 & 3.3 & 1.29 \\ \midrule

KIOU & \checkmark &  & 1 & 62.7 & 91.4 & 91.0 & 91.6 & 100.0 & 96.6 & 100.0 & 85.4 & 0.9 & 0.71 \\
KIOU & \checkmark &  & 2 & 51.4 & 82.6 & 92.0 & 83.3 & 99.4 & 86.0 & 100.0 & 57.0 & 4.4 & 1.09 \\
KIOU & \checkmark &  & 3 & 39.3 & 76.7 & 91.7 & 77.0 & 100.0 & 93.2 & 100.0 & 50.9 & 6.0 & 2.10 \\
KIOU & \checkmark &  & \textit{avg} & 51.1 & 83.6 & \textbf{91.6} & 83.9 & \textbf{99.8 }& 91.9 & \textbf{100.0} & 64.4 & 3.8 & 1.30 \\ \midrule

Byte &  &  & 1 & 23.6 & 73.5 & 90.1 & 91.1 & 85.4 & 96.6 & 97.6 & 85.1 & 0.9 & 9.40 \\
Byte &  &  & 2 & 28.2 & 68.5 & 91.6 & 83.7 & 85.3 & 86.0 & 86.8 & 60.5 & 3.5 & 4.93 \\
Byte &  &  & 3 & 24.9 & 67.9 & 91.3 & 79.3 & 88.1 & 93.6 & 89.9 & 54.1 & 4.3 & 6.33 \\
Byte &  &  & \textit{avg} & 25.6 & 70.0 & 91.0 & 84.7 & 86.3 & \textbf{92.0} & 91.4 & 66.6 & \textbf{2.9} & 6.88 \\ \midrule

KIOU &  &  & 1 & 23.6 & 73.5 & 90.1 & 91.1 & 85.4 & 96.6 & 97.6 & 85.1 & 0.9 & 9.40 \\
KIOU &  &  & 2 & 28.2 & 68.5 & 91.6 & 83.7 & 85.3 & 86.0 & 86.8 & 60.5 & 3.5 & 4.93 \\
KIOU &  &  & 3 & 24.9 & 67.9 & 91.3 & 79.3 & 88.1 & 93.6 & 89.9 & 54.1 & 4.3 & 6.33 \\
KIOU &  &  & \textit{avg} & 25.6 & 70.0 & 91.0 & 84.7 & 86.3 & \textbf{92.0} & 91.4 & 66.6 & \textbf{2.9 }& 6.88 \\ \bottomrule

\end{tabular}
\caption{Results for all tracking pipelines using ground truth (GT) detections on each scene. Results include higher order tracking accuracy (HOTA), multiple object tracking accuracy / precision (MOTA/MOTP), recall (Rec), precision (Prec), ground truth and prediction match rates (GT\% / Pred \%), mostly tracked and mostly lost objects (MT/ML) and number of ID switches per ground-truth object (Sw/GT). Best average result for each metric across all pipelines shown in bold. }
\label{tab:GT-detections}
\end{table}

\begin{table}[H]
\setlength{\tabcolsep}{1.5pt} % Default value: 6pt
\renewcommand{\arraystretch}{0.8}

\centering
\begin{tabular}{l|l|cccccccccc} \toprule
\textbf{TF} & \textbf{Scene} & \textbf{HOTA} & \textbf{MOTA} & \textbf{MOTP} & \textbf{Rec} & \textbf{Prec} & \textbf{GT\%} & \textbf{Pred\%} & \textbf{MT} & \textbf{ML} & \textbf{Sw/GT} \\ \toprule
\checkmark & 1 & 59.5 & 76.2 & 78.5 & 88.2 & 88.1 & 100.0 & 99.1 & 82.6 & 0.9 & 0.16 \\
\checkmark & 2 & 77.6 & 95.6 & 88.3 & 97.0 & 98.6 & 100.0 & 98.5 & 95.6 & 2.6 & 0.22 \\
\checkmark & 3 & 47.9 & 70.6 & 84.1 & 95.7 & 79.3 & 99.3 & 92.1 & 96.4 & 1.1 & 1.16 \\
\checkmark & \textit{avg} & \textbf{61.7} & \textbf{80.8} & 83.6 & 93.6 & \textbf{88.7} & 99.8 & \textbf{96.6} & 91.6 & 1.5 & \textbf{0.51} \\ \midrule

 & 1 & 19.9 & 14.6 & 90.3 & 89.2 & 55.0 & 100.0 & 76.9 & 84.2 & 0.3 & 9.15 \\
 & 2 & 29.3 & 38.7 & 92.4 & 95.1 & 63.0 & 100.0 & 83.0 & 98.2 & 0.0 & 5.25 \\
 & 3 & 28.7 & 36.9 & 92.8 & 98.2 & 61.8 & 100.0 & 83.5 & 98.6 & 0.0 & 5.25 \\
& \textit{avg} & 26.0 & 30.1 & \textbf{91.8} & \textbf{94.2} & 59.9 & \textbf{100.0} & 81.1 & \textbf{93.7} & \textbf{0.1} & 6.55 \\ \bottomrule

\end{tabular}
\caption{Results for all tracking pipelines using ground truth single camera tracklets on each scene. Best average result for each metric across all pipelines shown in bold.}
\label{tab:GT-tracklets}
\end{table}

\subsection*{Per-scene Results for Tracking Pipelines}

Tables \ref{tab:Dual3D}, \ref{tab:Single3D}, and \ref{tab:Crop3D} report the results for each pipeline using the Dual3D, Single3D, and Crop3D detectors, respectively, on each scene.  Across most pipelines, Scene 3 is the most difficult and Scene 1 is the easiest. On Scene 1, the best performing pipeline (Dual3D + KIOU + TF) achievs 58.5\% HOTA and 86.3\% mostly tracked objects, still not accurate enough for fine-grained traffic analyses (HOTA 0.75 and 95\% mostly tracked objects). The best-performing pipelines for Scene 3 (Single3D or Dual3D + KIOU + TF) acheive just 29.1\% HOTA).  All pipelines utilizing Crop3D perform poorly on Scene 2 (best HOTA 11.5\%). This is because Crop3D searches within a local region around each object prior, and always utilizes the best detection from this local crop to update the object's position. This strategy fails when the region is occluded (e.g. by snow) which is often the case in Scene 2. 

Finally, Figure \ref{fig:HOTA} shows the HOTA curves for the best performing pipeline (Dual3D + KIOU + TF) relative to the baseline with no cross-camera rectification, for Scene 1. \textit{Detection} and \textit{Association} scores refers to the DetA formula and AssA defined in \cite{luiten2021hota}, which are roughly meant to appraise the accuracy of the detection and object matching performance of the tracker independently. \textit{Higher Order Tracking Accuracy} (HOTA) is an aggregate metric composed of these two components, evaluated at 19 evenly spaced IOU thresholds required for a prediction to be considered a true positive. Lower thresholds result in higher scores because more predictions are deemed valid matches according to the threshold. Note that for both pipelines, the association score is lower than the detection score, indicating that more the cross-camera association problem is more problematic than detection accuracy for achieving high HOTA. Additionally, it can be seen that performance of all metrics declines steeply at a required IOU of 0.5 and higher, meaning that more precise object localization could likely also improve HOTA scores considerably.

\begin{figure}[H]
    \centering
    \includegraphics[width=0.7\textwidth]{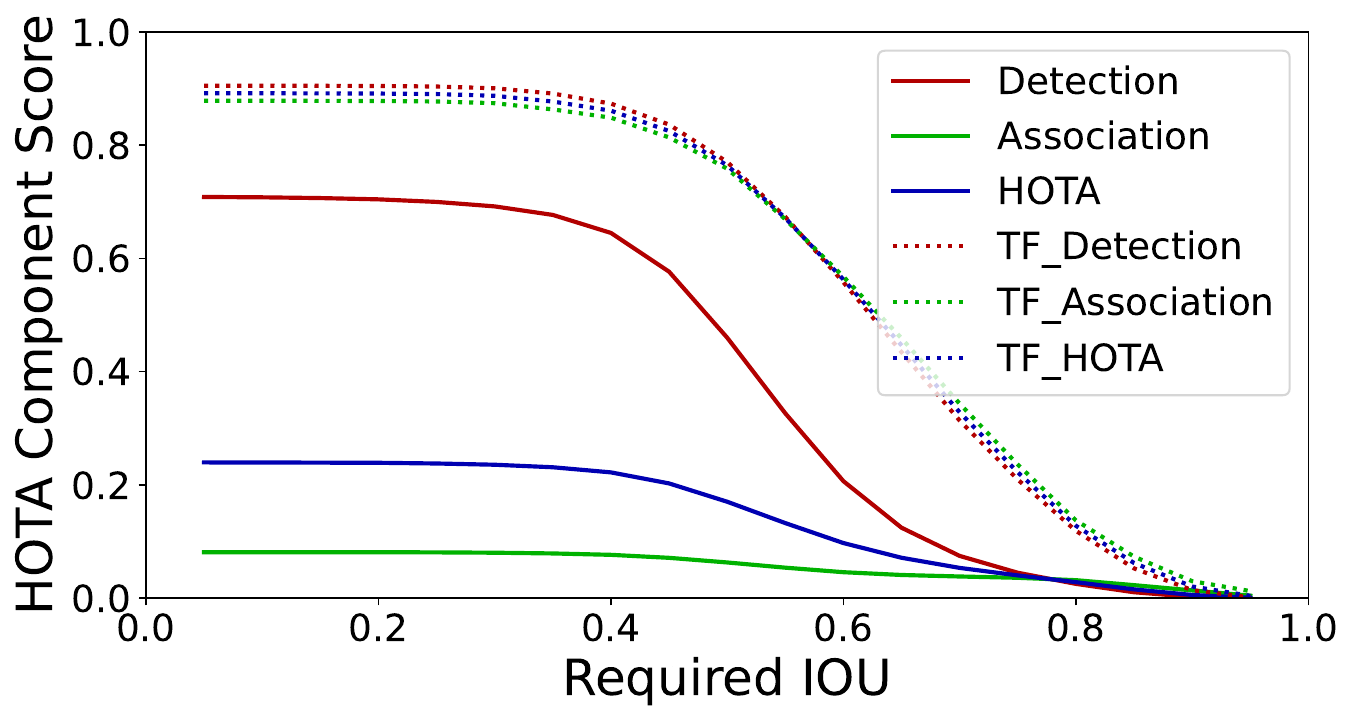}
    \caption{\textit{Detection} accuracy, \textit{Association} accuracy, and HOTA as defined for Dual3D + KIOU pipeline (Solid) and Dual3D + KIOU + TF (dotted).}
    \label{fig:HOTA}
\end{figure}

\begin{table}[H]
\setlength{\tabcolsep}{1.5pt} % Default value: 6pt
\renewcommand{\arraystretch}{0.8}
\centering
\begin{tabular}{lcc|l|cccccccccc} \toprule
\textbf{Tra.} & \textbf{DF} & \textbf{TF} & \textbf{Scene} & \textbf{HOTA} & \textbf{MOTA} & \textbf{MOTP} & \textbf{Rec} & \textbf{Prec} & \textbf{GT\%} & \textbf{Pred\%} & \textbf{MT} & \textbf{ML} & \textbf{Sw/GT} \\ \toprule

Byte & \checkmark & \checkmark & 1 & 25.9 & 19.2 & 49.0 & 54.1 & 60.8 & 88.5 & 91.2 & 26.4 & 25.2 & 0.17 \\
Byte & \checkmark & \checkmark & 2 & 43.2 & 74.8 & 73.9 & 80.8 & 93.2 & 91.2 & 89.7 & 55.3 & 8.8 & 0.57 \\
Byte & \checkmark & \checkmark & 3 & 23.6 & 56.0 & 68.3 & 61.9 & 91.5 & 92.2 & 99.3 & 26.0 & 11.0 & 1.98 \\
Byte & \checkmark & \checkmark & \textit{avg} & 30.9 & 50.0 & 63.7 & 65.6 & 81.9 & 90.6 & 93.4 & 35.9 & 15.0 & 0.91 \\ \midrule

KIOU & \checkmark & \checkmark & 1 & 52.1 & 86.2 & 68.9 & 89.3 & 96.6 & 95.0 & 99.5 & 80.1 & 3.1 & 0.19 \\
KIOU & \checkmark & \checkmark & 2 & 43.1 & 74.5 & 74.0 & 80.1 & 93.6 & 91.2 & 90.3 & 53.5 & 8.8 & 0.57 \\
KIOU & \checkmark & \checkmark & 3 & 24.0 & 54.0 & 64.5 & 60.2 & 90.9 & 88.3 & 96.1 & 23.8 & 19.2 & 1.29 \\
KIOU & \checkmark & \checkmark & \textit{avg} & 39.7 & 71.6 & 69.2 & 76.5 & 93.7 & 91.5 & \textbf{95.3} & 52.5 & 10.4 & 0.68 \\ \midrule

Byte &  & \checkmark & 1 & 57.6 & 89.0 & 69.1 & 92.2 & 96.6 & 95.3 & 98.8 & 84.2 & 2.5 & 0.04 \\
Byte &  & \checkmark & 2 & 29.8 & 73.1 & 62.8 & 78.8 & 93.4 & 93.9 & 86.1 & 52.6 & 7.0 & 1.30 \\
Byte &  & \checkmark & 3 & 28.7 & 62.9 & 64.9 & 69.5 & 91.4 & 89.7 & 96.0 & 40.2 & 14.6 & 1.07 \\
Byte &  & \checkmark & \textit{avg} & 38.7 & 75.0 & 65.6 & 80.2 & 93.8 & 93.0 & 93.6 & 59.0 & 8.0 & 0.80 \\  \midrule

KIOU &  & \checkmark & 1 & 58.5 & 89.7 & 69.2 & 92.9 & 96.7 & 95.3 & 98.4 & 86.3 & 2.2 & 0.02 \\
KIOU &  & \checkmark & 2 & 46.9 & 77.7 & 74.5 & 86.2 & 91.1 & 90.4 & 82.4 & 64.0 & 9.6 & 0.49 \\
KIOU &  & \checkmark & 3 & 29.1 & 63.5 & 64.8 & 69.9 & 91.7 & 89.3 & 96.1 & 40.9 & 14.6 & 1.05 \\
KIOU &  & \checkmark & \textit{avg} & \textbf{44.8} & \textbf{77.0} & \textbf{69.5} & \textbf{83.0} & 93.2 & 91.7 & 92.3 & \textbf{63.8} & 8.8 & \textbf{0.52} \\  \midrule

Byte & \checkmark &  & 1 & 26.5 & 72.8 & 57.8 & 77.2 & 95.0 & 94.7 & 97.9 & 57.8 & 4.0 & 1.51 \\
Byte & \checkmark &  & 2 & 18.7 & 53.2 & 70.3 & 59.0 & 91.7 & 91.2 & 87.3 & 14.0 & 9.6 & 3.50 \\
Byte & \checkmark &  & 3 & 17.5 & 54.6 & 66.5 & 56.2 & 97.7 & 92.2 & 98.7 & 16.7 & 12.5 & 3.27 \\
Byte & \checkmark &  & \textit{avg} & 20.9 & 60.2 & 64.9 & 64.2 & 94.8 & 92.7 & 94.7 & 29.5 & 8.7 & 2.76 \\ \midrule

KIOU & \checkmark &  & 1 & 26.8 & 73.4 & 57.9 & 77.3 & 95.6 & 94.4 & 98.5 & 57.1 & 4.0 & 1.50 \\
KIOU & \checkmark &  & 2 & 19.1 & 53.3 & 70.5 & 58.6 & 92.3 & 91.2 & 86.9 & 14.9 & 10.5 & 3.35 \\
KIOU & \checkmark &  & 3 & 17.5 & 54.5 & 66.6 & 56.2 & 97.7 & 92.2 & 98.8 & 17.1 & 12.1 & 3.30 \\
KIOU & \checkmark &  & \textit{avg} & 21.1 & 60.4 & 65.0 & 64.0 & \textbf{95.2} & 92.6 & 94.7 & 29.7 & 8.9 & 2.72 \\ \midrule

Byte &  &  & 1 & 14.1 & 60.5 & 57.9 & 82.1 & 80.7 & 94.7 & 94.7 & 66.5 & 3.1 & 8.57 \\
Byte &  &  & 2 & 15.6 & 50.1 & 69.4 & 68.6 & 79.5 & 91.2 & 79.1 & 25.4 & 8.8 & 6.66 \\
Byte &  &  & 3 & 15.2 & 54.8 & 66.8 & 67.6 & 84.8 & \textbf{93.6} & 88.7 & 35.6 & 8.5 & 6.57 \\
Byte &  &  & \textit{avg} & 15.0 & 55.1 & 64.7 & 72.8 & 81.7 & 93.2 & 87.5 & 42.5 & \textbf{6.8} & 7.27 \\ \midrule

KIOU &  &  & 1 & 14.2 & 61.3 & 58.1 & 82.5 & 81.0 & 94.7 & 94.9 & 67.4 & 3.4 & 8.63 \\
KIOU &  &  & 2 & 15.7 & 50.8 & 69.7 & 67.9 & 80.6 & 91.2 & 79.3 & 22.8 & 8.8 & 6.55 \\
KIOU &  &  & 3 & 15.2 & 54.8 & 66.9 & 67.6 & 84.9 & 93.2 & 89.0 & 36.7 & 8.9 & 6.57 \\
KIOU &  &  & \textit{avg} & 15.1 & 55.6 & 64.9 & 72.7 & 82.2 & 93.1 & 87.8 & 42.3 & 7.0 & 7.25 \\ \bottomrule

\end{tabular}
\caption{Results for all tracking pipelines using Dual3D detections on each scene. Best average result for each metric across all pipelines shown in bold.}
\label{tab:Dual3D}
\end{table}

\begin{table}[H]
\setlength{\tabcolsep}{1.5pt} % Default value: 6pt
\renewcommand{\arraystretch}{0.8}
\centering
\begin{tabular}{lcc|l|cccccccccc} \toprule
\textbf{Tra.} & \textbf{DF} & \textbf{TF} & \textbf{Scene} & \textbf{HOTA} & \textbf{MOTA} & \textbf{MOTP} & \textbf{Rec} & \textbf{Prec} & \textbf{GT\%} & \textbf{Pred\%} & \textbf{MT} & \textbf{ML} & \textbf{Sw/GT} \\ \toprule

Byte & \checkmark & \checkmark & 1 & 28.1 & 26.6 & 51.5 & 58.0 & 64.9 & 90.7 & 91.7 & 33.9 & 23.6 & 0.19 \\
Byte & \checkmark & \checkmark & 2 & 29.9 & 66.8 & 65.8 & 70.9 & 94.6 & 95.6 & 86.7 & 34.2 & 5.3 & 1.16 \\
Byte & \checkmark & \checkmark & 3 & 24.4 & 54.5 & 65.2 & 59.6 & 92.3 & 90.0 & 96.9 & 21.0 & 17.4 & 1.25 \\
Byte & \checkmark & \checkmark & \textit{avg} & 27.5 & 49.3 & 60.8 & 62.8 & 83.9 & 92.1 & 91.8 & 29.7 & 15.4 & 0.86 \\ \midrule

KIOU & \checkmark & \checkmark & 1 & 52.3 & 86.1 & 69.1 & 89.4 & 96.5 & 96.3 & 99.5 & 81.7 & 1.9 & 0.17 \\
KIOU & \checkmark & \checkmark & 2 & 42.9 & 74.1 & 75.2 & 79.7 & 93.5 & 93.9 & 91.2 & 51.8 & 7.0 & 0.61 \\
KIOU & \checkmark & \checkmark & 3 & 24.5 & 54.5 & 65.2 & 59.7 & 92.2 & 90.0 & 96.3 & 21.4 & 17.4 & 1.24 \\
KIOU & \checkmark & \checkmark & \textit{avg} & 39.9 & 71.6 & 69.8 & 76.3 & 94.1 & 93.4 & \textbf{95.6} & 51.6 & 8.8 & 0.67 \\ \midrule

Byte &  & \checkmark & 1 & 57.8 & 89.2 & 69.2 & 92.7 & 96.4 & 96.6 & 98.5 & 86.6 & 1.6 & 0.02 \\
Byte &  & \checkmark & 2 & 31.9 & 73.5 & 65.2 & 78.8 & 93.8 & 95.6 & 85.5 & 50.0 & 4.4 & 1.22 \\
Byte &  & \checkmark & 3 & 25.0 & 55.1 & 70.0 & 70.3 & 82.5 & 92.5 & 93.5 & 39.5 & 8.2 & 2.19 \\
Byte &  & \checkmark & \textit{avg} & 38.2 & 72.6 & 68.2 & 80.6 & 90.9 & \textbf{94.9} & 92.5 & 58.7 & \textbf{4.7} & 1.15 \\ \midrule

KIOU &  & \checkmark & 1 & 57.5 & 89.3 & 69.2 & 92.9 & 96.3 & 96.6 & 97.9 & 87.0 & 1.6 & 0.03 \\
KIOU &  & \checkmark & 2 & 47.7 & 78.2 & 75.4 & 86.4 & 91.5 & 93.0 & 79.8 & 64.0 & 7.0 & 0.47 \\
KIOU &  & \checkmark & 3 & 29.1 & 63.8 & 65.1 & 69.7 & 92.3 & 90.4 & 96.1 & 35.6 & 14.9 & 1.05 \\
KIOU &  & \checkmark & \textit{avg} & \textbf{44.8} & \textbf{77.1} & \textbf{69.9} & \textbf{83.0} & 93.4 & 93.3 & 91.3 & \textbf{62.2} & 7.8 & \textbf{0.52} \\ \midrule

Byte & \checkmark &  & 1 & 26.1 & 72.5 & 58.2 & 76.6 & 95.2 & 96.0 & 97.8 & 55.3 & 2.5 & 1.55 \\
Byte & \checkmark &  & 2 & 19.3 & 54.2 & 70.9 & 59.0 & 93.0 & 93.9 & 86.0 & 11.4 & 7.9 & 3.31 \\
Byte & \checkmark &  & 3 & 18.4 & 54.3 & 69.2 & 56.0 & 97.6 & 91.5 & 98.7 & 14.6 & 12.1 & 3.09 \\
Byte & \checkmark &  & \textit{avg} & 21.3 & 60.3 & 66.1 & 63.9 & 95.3 & 93.8 & 94.2 & 27.1 & 7.5 & 2.65 \\ \midrule

KIOU & \checkmark &  & 1 & 26.4 & 72.6 & 58.4 & 76.6 & 95.5 & 96.3 & 97.7 & 53.7 & 2.2 & 1.52 \\
KIOU & \checkmark &  & 2 & 19.4 & 53.9 & 71.2 & 58.5 & 93.2 & 94.7 & 85.5 & 11.4 & 7.9 & 3.25 \\
KIOU & \checkmark &  & 3 & 18.4 & 54.3 & 69.2 & 55.9 & 97.7 & 91.5 & 98.4 & 14.2 & 12.5 & 3.09 \\
KIOU & \checkmark &  & \textit{avg} & 21.4 & 60.3 & 66.2 & 63.7 & \textbf{95.5} & 94.2 & 93.9 & 26.5 & 7.5 & 2.62 \\ \midrule

Byte &  &  & 1 & 14.0 & 59.9 & 57.9 & 81.6 & 80.5 & 96.6 & 94.1 & 64.6 & 1.9 & 8.51 \\
Byte &  &  & 2 & 15.9 & 50.9 & 69.8 & 68.8 & 80.1 & 93.9 & 76.7 & 24.6 & 6.1 & 6.48 \\
Byte &  &  & 3 & 15.4 & 51.1 & 69.0 & 66.5 & 81.9 & 92.5 & 87.0 & 32.4 & 9.3 & 6.49 \\
Byte &  &  & \textit{avg} & 15.1 & 54.0 & 65.5 & 72.3 & 80.8 & 94.3 & 85.9 & 40.5 & 5.8 & 7.16 \\ \midrule

KIOU &  &  & 1 & 14.2 & 61.0 & 58.1 & 82.2 & 81.0 & 96.6 & 94.4 & 65.5 & 1.6 & 8.53 \\
KIOU &  &  & 2 & 16.0 & 50.8 & 70.1 & 67.8 & 80.6 & 94.7 & 77.0 & 21.1 & 6.1 & 6.32 \\
KIOU &  &  & 3 & 15.4 & 51.4 & 69.0 & 66.5 & 82.2 & 92.2 & 87.1 & 31.7 & 9.3 & 6.49 \\
KIOU &  &  & \textit{avg} & 15.2 & 54.4 & 65.7 & 72.2 & 81.3 & 94.5 & 86.1 & 39.4 & 5.6 & 7.12 \\ \bottomrule

\end{tabular}
\caption{Results for all tracking pipelines using Single3D detections on each scene. Best average result for each metric across all pipelines shown in bold.}
\label{tab:Single3D}
\end{table}

\begin{table}[H]
\setlength{\tabcolsep}{1.5pt} % Default value: 6pt
\renewcommand{\arraystretch}{0.8}
\centering
\begin{tabular}{lcc|l|cccccccccc} \toprule
\textbf{Tra.} & \textbf{DF} & \textbf{TF} & \textbf{Scene} & \textbf{HOTA} & \textbf{MOTA} & \textbf{MOTP} & \textbf{Rec} & \textbf{Prec} & \textbf{GT\%} & \textbf{Pred\%} & \textbf{MT} & \textbf{ML} & \textbf{Sw/GT} \\ \toprule

Byte & \checkmark & \checkmark & 1 & 35.8 & 43.5 & 68.4 & 68.7 & 73.3 & 92.5 & 82.3 & 49.7 & 13.4 & 0.34 \\
Byte & \checkmark & \checkmark & 2 & 11.4 & -27.9 & 66.3 & 31.7 & 34.8 & 90.4 & 44.6 & 6.1 & 40.4 & 1.89 \\
Byte & \checkmark & \checkmark & 3 & 23.5 & 48.2 & 65.3 & 59.8 & 84.0 & 88.6 & 91.8 & 21.0 & 21.4 & 1.20 \\
Byte & \checkmark & \checkmark & \textit{avg} & 23.6 & 21.3 & 66.7 & 53.4 & 64.0 & 90.5 & 72.9 & 25.6 & 25.0 & \textbf{1.14} \\ \midrule

KIOU & \checkmark & \checkmark & 1 & 36.7 & 46.5 & 68.7 & 71.0 & 74.4 & 93.2 & 81.7 & 50.0 & 10.6 & 0.35 \\
KIOU & \checkmark & \checkmark & 2 & 11.5 & -34.0 & 68.2 & 30.5 & 32.2 & 90.4 & 38.2 & 4.4 & 39.5 & 1.70 \\
KIOU & \checkmark & \checkmark & 3 & 25.6 & 51.8 & 69.3 & 61.9 & 86.2 & 87.9 & 93.8 & 28.5 & 16.7 & 1.36 \\
KIOU & \checkmark & \checkmark & \textit{avg} & \textbf{24.6} & 21.4 & \textbf{68.7} & 54.4 & 64.2 & 90.5 & 71.2 & 27.6 & 22.3 & \textbf{1.14} \\ \midrule

Byte &  & \checkmark & 1 & 13.2 & -31.7 & 47.7 & 38.8 & 35.6 & 89.4 & 59.5 & 14.0 & 35.4 & 0.86 \\
Byte &  & \checkmark & 2 & 9.1 & -65.8 & 66.2 & 29.7 & 23.8 & 91.2 & 34.3 & 6.1 & 39.5 & 2.30 \\
Byte &  & \checkmark & 3 & 23.3 & 47.9 & 65.5 & 62.1 & 81.6 & 90.4 & 85.5 & 23.5 & 21.7 & 1.36 \\
Byte &  & \checkmark & \textit{avg} & 15.2 & -16.5 & 59.8 & 43.5 & 47.0 & 90.4 & 59.8 & 14.5 & 32.2 & 1.51 \\ \midrule

KIOU &  & \checkmark & 1 & 32.4 & 26.7 & 67.6 & 67.8 & 62.3 & 93.2 & 62.8 & 48.8 & 14.6 & 0.43 \\
KIOU &  & \checkmark & 2 & 7.1 & -74.0 & 61.3 & 26.2 & 20.8 & 89.5 & 26.9 & 3.5 & 41.2 & 2.67 \\
KIOU &  & \checkmark & 3 & 22.8 & 40.9 & 69.4 & 60.8 & 75.4 & 87.9 & 71.5 & 24.2 & 23.5 & 1.53 \\
KIOU &  & \checkmark & \textit{avg} & 20.7 & -2.1 & 66.1 & 51.6 & 52.9 & 90.2 & 53.7 & 25.5 & 26.4 & 1.54 \\ \midrule

Byte & \checkmark &  & 1 & 24.7 & 58.4 & 59.0 & 74.7 & 82.4 & 94.4 & 93.5 & 51.9 & 5.3 & 1.43 \\
Byte & \checkmark &  & 2 & 10.9 & -11.2 & 66.7 & 37.8 & 43.8 & 92.1 & 53.1 & 3.5 & 28.9 & 3.55 \\
Byte & \checkmark &  & 3 & 22.0 & 56.9 & 68.7 & 62.4 & 92.2 & 89.3 & 97.7 & 26.0 & 15.7 & 2.22 \\
Byte & \checkmark &  & \textit{avg} & 19.2 & \textbf{34.7} & 64.8 & 58.3 & \textbf{72.8} & \textbf{91.9} & \textbf{81.4} & 27.1 & 16.6 & 2.40 \\ \midrule

KIOU & \checkmark &  & 1 & 25.2 & 59.0 & 59.1 & 75.6 & 82.3 & 95.0 & 94.1 & 51.2 & 4.3 & 1.42 \\
KIOU & \checkmark &  & 2 & 10.5 & -20.5 & 67.0 & 35.4 & 38.9 & 92.1 & 47.6 & 1.8 & 32.5 & 3.49 \\
KIOU & \checkmark &  & 3 & 22.2 & 57.7 & 68.8 & 62.8 & 92.9 & 88.6 & 97.3 & 24.6 & 14.9 & 2.21 \\
KIOU & \checkmark &  & \textit{avg} & 19.3 & 32.1 & 65.0 & 57.9 & 71.4 & \textbf{91.9} & 79.7 & 25.9 & 17.3 & 2.37 \\ \midrule

Byte &  &  & 1 & 22.0 & 45.1 & 59.0 & 76.4 & 71.2 & 94.1 & 81.9 & 56.8 & 5.0 & 1.90 \\
Byte &  &  & 2 & 9.2 & -46.2 & 66.1 & 37.9 & 31.2 & 91.2 & 42.8 & 4.4 & 24.6 & 4.62 \\
Byte &  &  & 3 & 21.4 & 57.4 & 68.7 & 65.1 & 89.7 & 90.4 & 94.2 & 29.9 & 15.7 & 2.61 \\
Byte &  &  & \textit{avg} & 17.6 & 18.7 & 64.6 & \textbf{59.8} & 64.0 & \textbf{91.9} & 73.0 & \textbf{30.4} & \textbf{15.1} & 3.04 \\ \midrule

KIOU &  &  & 1 & 22.5 & 47.8 & 59.2 & 77.1 & 72.8 & 95.0 & 83.0 & 58.4 & 4.3 & 1.89 \\
KIOU &  &  & 2 & 8.2 & -62.8 & 66.0 & 33.4 & 25.9 & 90.4 & 39.7 & 1.8 & 28.9 & 5.09 \\
KIOU &  &  & 3 & 20.0 & 47.5 & 69.4 & 62.0 & 81.3 & 90.4 & 75.5 & 24.6 & 22.1 & 2.74 \\
KIOU &  &  & \textit{avg} & 16.9 & 10.8 & 64.9 & 57.5 & 60.0 & \textbf{91.9} & 66.0 & 28.2 & 18.5 & 3.24 \\  \midrule

\end{tabular}
\caption{Results for all tracking pipelines using Crop3D detections on each scene. Best average result for each metric across all pipelines shown in bold.}
\label{tab:Crop3D}
\end{table}

\section*{Appendix VII: Privacy Considerations}
As with any dataset containing video data of a public location, the I24-3D dataset potentially contains \textit{personally identifiable information} (PII). We visually inspect video sequences to ensure that license plates are not visible at a visually distinctive level (license plate numbers cannot be determined from imagery except possibly with extensive de-noising techniques). We further process each video sequence with a license plate blurring software. Likewise, we confirm that driver faces in each vehicle are not visually discernible, no pedestrians are visible within the dataset, and no anomalous events (e.g. crashes) occur. Lastly, we have submitted this research to University \textit{Institutional Review Board} (IRB) and secured research approval to ensure that the dataset management protocols appropriately protect individuals' privacy.

\bibliography{sources}